\documentclass[10pt,twocolumn,letterpaper]{article}

\usepackage[pagenumbers]{cvpr}  

\usepackage{booktabs}
\usepackage{multirow}
\usepackage{longtable}
\usepackage{placeins}
\newcommand{\tableuncert}{\footnotesize}
\newcommand{\tablebody}{\small}

\usepackage{microtype}

\input{numbers}
\expandafter\def\csname gp@e49native\endcsname{+11.4}
\expandafter\def\csname gpci@e49native\endcsname{[+9.1, +13.8]}
\expandafter\def\csname gp@e49resolutionReleased\endcsname{+1.9}
\expandafter\def\csname gpci@e49resolutionReleased\endcsname{[+0.8, +3.0]}
\expandafter\def\csname gp@e49resolutionAdapted\endcsname{$-$0.1}
\expandafter\def\csname gpci@e49resolutionAdapted\endcsname{[$-$1.7, +1.6]}
\expandafter\def\csname nm@e49Wins\endcsname{13}
\expandafter\def\csname nm@e49Comparisons\endcsname{13}

\expandafter\def\csname nm@breakEvenBaseFirstKlein\endcsname{399,000}
\expandafter\def\csname nm@breakEvenBaseFirstSD\endcsname{807,000}

\expandafter\def\csname nm@e50PoolTemplateKleinVsReleased\endcsname{+2.1}
\expandafter\def\csname nm@e50PoolTemplateKleinVsReleasedCI\endcsname{[+0.2, +4.0]}
\expandafter\def\csname nm@e50PoolTemplateSDVsReleased\endcsname{+16.7}
\expandafter\def\csname nm@e50PoolTemplateSDVsReleasedCI\endcsname{[+14.1, +19.4]}
\expandafter\def\csname nm@e50PoolLlmKleinVsReleased\endcsname{+5.4}
\expandafter\def\csname nm@e50PoolLlmKleinVsReleasedCI\endcsname{[+3.4, +7.3]}
\expandafter\def\csname nm@e50PoolLlmSDVsReleased\endcsname{+15.6}
\expandafter\def\csname nm@e50PoolLlmSDVsReleasedCI\endcsname{[+12.9, +18.5]}
\expandafter\def\csname nm@e50OriginalKleinVsLlmReleased\endcsname{+2.7}
\expandafter\def\csname nm@e50OriginalKleinVsLlmReleasedCI\endcsname{[+2.0, +3.4]}
\expandafter\def\csname nm@e50OriginalSDVsLlmReleased\endcsname{+7.1}
\expandafter\def\csname nm@e50OriginalSDVsLlmReleasedCI\endcsname{[+6.3, +7.9]}
\expandafter\def\csname nm@e50ScaleKleinVsReleased\endcsname{$-$1.5}
\expandafter\def\csname nm@e50ScaleKleinVsReleasedCI\endcsname{[$-$2.0, $-$1.0]}
\expandafter\def\csname nm@e50ScaleSDVsReleased\endcsname{+6.9}
\expandafter\def\csname nm@e50ScaleSDVsReleasedCI\endcsname{[+6.2, +7.7]}
\expandafter\def\csname nm@e50ScaleKleinVsBaseFirst\endcsname{$-$2.4}
\expandafter\def\csname nm@e50ScaleKleinVsBaseFirstCI\endcsname{[$-$2.8, $-$2.0]}
\expandafter\def\csname nm@e50ScaleSDVsBaseFirst\endcsname{+5.2}
\expandafter\def\csname nm@e50ScaleSDVsBaseFirstCI\endcsname{[+4.8, +5.6]}
\expandafter\def\csname nm@e50OriginalKleinVsLlmBaseFirst\endcsname{+1.8}
\expandafter\def\csname nm@e50OriginalKleinVsLlmBaseFirstCI\endcsname{[+1.2, +2.4]}
\expandafter\def\csname nm@e50OriginalSDVsLlmBaseFirst\endcsname{+5.4}
\expandafter\def\csname nm@e50OriginalSDVsLlmBaseFirstCI\endcsname{[+4.8, +6.0]}
\expandafter\def\csname nm@e50FitKlein\endcsname{+3.3}
\expandafter\def\csname nm@e50FitKleinCI\endcsname{[+2.4, +4.2]}
\expandafter\def\csname nm@e50FitSD\endcsname{$-$1.1}
\expandafter\def\csname nm@e50FitSDCI\endcsname{[$-$1.6, $-$0.5]}
\expandafter\def\csname nm@e50LlmReleasedKlein\endcsname{+5.4}
\expandafter\def\csname nm@e50LlmReleasedKleinCI\endcsname{[+3.4, +7.3]}
\expandafter\def\csname nm@e50LlmReleasedSD\endcsname{+15.6}
\expandafter\def\csname nm@e50LlmReleasedSDCI\endcsname{[+12.9, +18.5]}
\expandafter\def\csname nm@e50LlmBaseFirstKlein\endcsname{+0.3}
\expandafter\def\csname nm@e50LlmBaseFirstKleinCI\endcsname{[$-$0.7, +1.3]}
\expandafter\def\csname nm@e50LlmBaseFirstSD\endcsname{+7.0}
\expandafter\def\csname nm@e50LlmBaseFirstSDCI\endcsname{[+5.6, +8.3]}
\expandafter\def\csname nm@e50TemplateReleasedKlein\endcsname{+2.1}
\expandafter\def\csname nm@e50TemplateReleasedKleinCI\endcsname{[+0.2, +4.0]}
\expandafter\def\csname nm@e50TemplateReleasedSD\endcsname{+16.7}
\expandafter\def\csname nm@e50TemplateReleasedSDCI\endcsname{[+14.1, +19.4]}
\expandafter\def\csname nm@e50TemplateBaseFirstKlein\endcsname{$-$3.0}
\expandafter\def\csname nm@e50TemplateBaseFirstKleinCI\endcsname{[$-$4.2, $-$1.9]}
\expandafter\def\csname nm@e50TemplateBaseFirstSD\endcsname{+8.0}
\expandafter\def\csname nm@e50TemplateBaseFirstSDCI\endcsname{[+6.6, +9.5]}
\expandafter\def\csname nm@e50TemplateEvalKlein\endcsname{$-$1.4}
\expandafter\def\csname nm@e50TemplateEvalKleinCI\endcsname{[$-$2.8, $-$0.1]}
\expandafter\def\csname nm@e50TemplateEvalSD\endcsname{$-$1.1}
\expandafter\def\csname nm@e50TemplateEvalSDCI\endcsname{[$-$2.0, $-$0.2]}
\expandafter\def\csname nm@e50Practicalin1k_llm_kb_dd2_g1Released\endcsname{+0.9}
\expandafter\def\csname nm@e50Practicalin1k_llm_kb_dd2_g1ReleasedCI\endcsname{[+0.4, +1.3]}
\expandafter\def\csname nm@e50Practicalin1k_kd_ad2Released\endcsname{+2.7}
\expandafter\def\csname nm@e50Practicalin1k_kd_ad2ReleasedCI\endcsname{[+2.0, +3.4]}
\expandafter\def\csname nm@e50Practicalin1k_kd_ad2BaseFirst\endcsname{+1.8}
\expandafter\def\csname nm@e50Practicalin1k_kd_ad2BaseFirstCI\endcsname{[+1.2, +2.4]}
\expandafter\def\csname nm@e50Practicalin1k_llm_kd_llmReleased\endcsname{$-$1.5}
\expandafter\def\csname nm@e50Practicalin1k_llm_kd_llmReleasedCI\endcsname{[$-$2.0, $-$1.0]}
\expandafter\def\csname nm@e50Practicalin1k_llm_kd_llmBaseFirst\endcsname{$-$2.4}
\expandafter\def\csname nm@e50Practicalin1k_llm_kd_llmBaseFirstCI\endcsname{[$-$2.8, $-$2.0]}
\expandafter\def\csname nm@e50Practicalin1k_llm_kd_ad2Released\endcsname{$-$3.3}
\expandafter\def\csname nm@e50Practicalin1k_llm_kd_ad2ReleasedCI\endcsname{[$-$3.9, $-$2.8]}
\expandafter\def\csname nm@e50Practicalin1k_llm_kd_ad2BaseFirst\endcsname{$-$4.2}
\expandafter\def\csname nm@e50Practicalin1k_llm_kd_ad2BaseFirstCI\endcsname{[$-$4.7, $-$3.8]}
\expandafter\def\csname nm@e50NineHundredkd_llm\endcsname{$-$1.6}
\expandafter\def\csname nm@e50NineHundredkd_llmCI\endcsname{[$-$2.1, $-$1.0]}
\expandafter\def\csname nm@e50Practicalin1k_llm_sdt4_dd2_g1Released\endcsname{+1.7}
\expandafter\def\csname nm@e50Practicalin1k_llm_sdt4_dd2_g1ReleasedCI\endcsname{[+1.1, +2.4]}
\expandafter\def\csname nm@e50Practicalin1k_sdt_ad1Released\endcsname{+7.1}
\expandafter\def\csname nm@e50Practicalin1k_sdt_ad1ReleasedCI\endcsname{[+6.3, +7.9]}
\expandafter\def\csname nm@e50Practicalin1k_sdt_ad1BaseFirst\endcsname{+5.4}
\expandafter\def\csname nm@e50Practicalin1k_sdt_ad1BaseFirstCI\endcsname{[+4.8, +6.0]}
\expandafter\def\csname nm@e50Practicalin1k_llm_sdt_ad1Released\endcsname{+6.9}
\expandafter\def\csname nm@e50Practicalin1k_llm_sdt_ad1ReleasedCI\endcsname{[+6.2, +7.7]}
\expandafter\def\csname nm@e50Practicalin1k_llm_sdt_ad1BaseFirst\endcsname{+5.2}
\expandafter\def\csname nm@e50Practicalin1k_llm_sdt_ad1BaseFirstCI\endcsname{[+4.8, +5.6]}
\expandafter\def\csname nm@e50NineHundredsdt_ad1\endcsname{+6.9}
\expandafter\def\csname nm@e50NineHundredsdt_ad1CI\endcsname{[+6.1, +7.7]}
\expandafter\def\csname nm@e50PostHocKleinVsReleased\endcsname{$-$3.3}
\expandafter\def\csname nm@e50PostHocKleinVsReleasedCI\endcsname{[$-$3.9, $-$2.8]}
\expandafter\def\csname nm@e50PostHocKleinVsBaseFirst\endcsname{$-$4.2}
\expandafter\def\csname nm@e50PostHocKleinVsBaseFirstCI\endcsname{[$-$4.7, $-$3.8]}
\expandafter\def\csname nm@e50LlmVsTemplateAtScale\endcsname{+1.9}
\expandafter\def\csname nm@e50LlmVsTemplateAtScaleCI\endcsname{[+1.6, +2.2]}
\expandafter\def\csname nm@e50PostHocKleinNineHundred\endcsname{$-$3.4}
\expandafter\def\csname nm@e50PostHocKleinNineHundredCI\endcsname{[$-$4.0, $-$2.9]}
\expandafter\def\csname nm@e50FitSDLoss\endcsname{1.1}
\expandafter\def\csname nm@e50PoorRun\endcsname{35.9}
\expandafter\def\csname nm@e50OtherRunOne\endcsname{43.3}
\expandafter\def\csname nm@e50OtherRunTwo\endcsname{42.5}
\expandafter\def\csname nm@e50Sensitivity\endcsname{$-$0.5}
\expandafter\def\csname nm@e50SensitivityCI\endcsname{[$-$1.5, +0.6]}

\input{numbers-review}   
\expandafter\def\csname tp@in1k_kb_dd2_g1_full\endcsname{30.8}
\expandafter\def\csname tpsd@in1k_kb_dd2_g1_full\endcsname{30.8$\pm$0.1}
\expandafter\def\csname tpn@in1k_kb_dd2_g1_full\endcsname{3}
\expandafter\def\csname tpnfe@in1k_kb_dd2_g1_full\endcsname{4}
\expandafter\def\csname tpsec@in1k_kb_dd2_g1_full\endcsname{0.153}
\expandafter\def\csname tp@in1k_sdt4_dd2_g1_full\endcsname{26.7}
\expandafter\def\csname tpsd@in1k_sdt4_dd2_g1_full\endcsname{26.7$\pm$0.3}
\expandafter\def\csname tpn@in1k_sdt4_dd2_g1_full\endcsname{3}
\expandafter\def\csname tpnfe@in1k_sdt4_dd2_g1_full\endcsname{4}
\expandafter\def\csname tpsec@in1k_sdt4_dd2_g1_full\endcsname{0.029}
\expandafter\def\csname gp@inkAdVsBfKlein\endcsname{+3.5}
\expandafter\def\csname gpci@inkAdVsBfKlein\endcsname{[+2.9, +4.1]}
\expandafter\def\csname gp@inkAdVsBfSD\endcsname{+5.4}
\expandafter\def\csname gpci@inkAdVsBfSD\endcsname{[+5.0, +5.8]}
\expandafter\def\csname nm@fullTotalBfKlein\endcsname{54.5}
\expandafter\def\csname nm@fullTotalAdKlein\endcsname{41.4}
\expandafter\def\csname nm@e51CrossoverMillionKlein\endcsname{0.4}
\expandafter\def\csname nm@fullTotalBfSD\endcsname{10.4}
\expandafter\def\csname nm@fullTotalAdSD\endcsname{10.2}
\expandafter\def\csname nm@e51CrossoverMillionSD\endcsname{1.1}

\expandafter\def\csname gp@e52KleinBaseVsReleased\endcsname{$-$2.4}
\expandafter\def\csname gpci@e52KleinBaseVsReleased\endcsname{[$-$3.0, $-$1.9]}
\expandafter\def\csname gp@e52KleinBaseVsBaseFirst\endcsname{$-$3.3}
\expandafter\def\csname gpci@e52KleinBaseVsBaseFirst\endcsname{[$-$3.8, $-$2.9]}
\expandafter\def\csname gp@e52SDBaseVsReleased\endcsname{+1.5}
\expandafter\def\csname gpci@e52SDBaseVsReleased\endcsname{[+0.8, +2.3]}
\expandafter\def\csname gp@e52SDBaseVsBaseFirst\endcsname{$-$0.2}
\expandafter\def\csname gpci@e52SDBaseVsBaseFirst\endcsname{[$-$0.8, +0.3]}
\expandafter\def\csname gp@e52KleinLlm100\endcsname{$-$4.6}
\expandafter\def\csname gpci@e52KleinLlm100\endcsname{[$-$6.2, $-$3.0]}
\expandafter\def\csname gp@e52KleinLlm500\endcsname{+5.2}
\expandafter\def\csname gpci@e52KleinLlm500\endcsname{[+3.0, +7.5]}
\expandafter\def\csname nm@e52KleinBaseLoss\endcsname{2.4}
\expandafter\def\csname nm@e52SDBaseGain\endcsname{1.5}

\expandafter\def\csname nm@inkFullShareSD\endcsname{62}

\definecolor{cvprblue}{rgb}{0.21,0.49,0.74}
\usepackage{xr-hyper}
\usepackage[pagebackref,breaklinks,colorlinks,allcolors=cvprblue]{hyperref}

\def\confName{CVPR}
\def\confYear{2027}

\title{Fast and Useful: Adapting Distilled Generators for Visual Learning}

\author{Patryk Bartkowiak\textsuperscript{1,4} \quad
Dominik Michels\textsuperscript{2,4} \quad
S{\"o}ren Pirk\textsuperscript{3,4} \quad
Wojciech Pa{\l}ubicki\textsuperscript{1,4}\\
\textsuperscript{1}Adam Mickiewicz University \qquad \textsuperscript{2}KAUST \qquad \textsuperscript{3}Kiel University \qquad \textsuperscript{4}GreenMatterAI
}

\makeatletter
\hypersetup{
  pdftitle={\@title},
  pdfsubject={Synthetic training data from adapted distilled text-to-image generators},
  pdfkeywords={synthetic data, visual learning, diffusion models, distillation, low-rank adaptation}
}
\makeatother
\iftoggle{cvprfinal}{%
  \hypersetup{pdfauthor={Patryk Bartkowiak and Dominik Michels and S{\"o}ren Pirk and Wojciech Pa{\l}ubicki}}%
}{%
  \hypersetup{pdfauthor={}}%
}

\makeatletter\patchcmd{\@maketitle}{\vskip .375in}{\vskip .15in}{}{\errmessage{patch failed}}\makeatother

\begin{document}
\twocolumn[{%
\renewcommand\twocolumn[1][]{#1}%
\maketitle
\vspace{-28pt}
\begin{center}
  \centering
  \captionsetup{type=figure}
  \includegraphics[width=\textwidth]{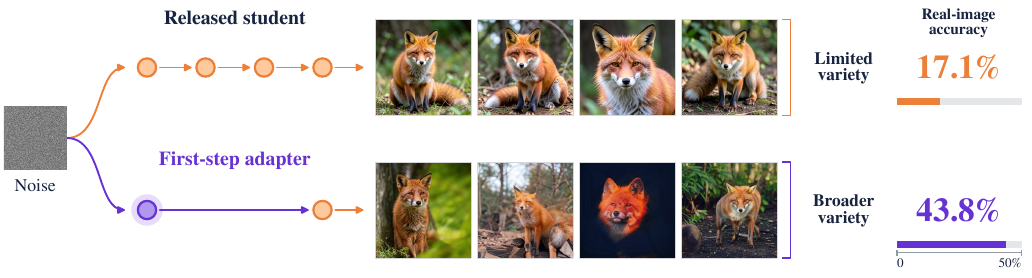}
  \captionof{figure}{\textbf{First-step adaptation restores variety and improves learning from synthetic images.}
  The adapter learns to jump to an intermediate state of the unguided base, replacing the released
  FLUX.2-klein student's first three predictions. The unchanged student then finishes generation,
  requiring two denoiser evaluations instead of four. Fox images show one selected class with matched
  prompts and noise seeds. ResNet-50 top-1 accuracy on real ImageNet-100 increases from \tpsd{klein}\%
  to \textbf{\tpsd{kd_ad2_x3}\%} after training on 500 generated images per class (mean $\pm$ SD).}
  \label{fig:teaser}
\end{center}

}]
\begin{abstract}
Distilled text-to-image models generate images quickly, but classifiers trained on those images can perform
substantially worse than classifiers trained on images from their slower base models. We adapt released
distilled generators to produce better training data while retaining fast generation. We fit a low-rank update
to the student's first prediction so that it reaches an intermediate state produced by its unguided base,
then disable the update and let the unchanged student finish generation. The base supplies training targets
and is absent at inference. Guided targets and shorter jumps give classifiers lower accuracy in our
ablations, showing why the state used for supervision matters. Across three generator families, the adapted
images train better classifiers than images from the released students. With class-name templates at full
ImageNet-1k scale, ResNet-50 top-1 rises from \tp{in1k_klein_full}\% to \tp{in1k_kd_ad2_full}\% for
FLUX.2-klein and from \tp{in1k_sdt4_full}\% to \tp{in1k_sdt_ad1_full}\% for SD-Turbo, using two and three
denoiser evaluations per image, respectively. Fixed adapters also improve results on additional classes,
fine-grained datasets, and other learner architectures and training recipes. Template-based adaptation remains
useful against richer-prompt alternatives, while the benefit of combining adaptation with LLM generation
prompts depends on the generator. Adapting a single generation step makes fast generators
substantially more useful for training visual models.
\end{abstract}

\section{Introduction}
\label{sec:intro}

Synthetic images can expand training sets for computer vision without requiring another round of image
collection and annotation. Text prompts provide control over the requested classes and allow datasets to be
generated at different scales, supporting both learning from synthetic images alone and supplementing limited
real data~\cite{sariyildiz2023fake,azizi2023synthetic,he2023synthetic}. Their usefulness depends on what visual
models learn from them: a generated dataset must help a classifier recognize real images.

Diffusion-based text-to-image models make this approach flexible, but generating every image through many
denoising steps becomes costly at dataset scale. Distillation compresses generation into a few predictions,
making large synthetic datasets faster to produce~\cite{sauer2024adversarial,yin2024improved}. That acceleration
can come with a loss in the usefulness of the generated images. Adamkiewicz \etal~\cite{adamkiewicz2026pretty}
show that classifiers trained on images from released distilled generators can perform worse than classifiers
trained on images from their full-step counterparts. We aim to retain the speed advantage while making the
released student's images more useful for learning.

We study this problem by comparing changes to the generator while holding the classifier training procedure
fixed. Lower guidance and richer prompts already improve synthetic-data learning~\cite{sariyildiz2023fake,
fan2024scaling}, and replacing a distilled model's first steps with its base can restore sample
diversity~\cite{gandikota2026distilling}. We measure how these choices affect classification on real images and
use the resulting comparisons to construct supervision for a fast student. Additional classifier architectures
and training recipes test whether the improvement extends beyond the primary learner.

Our approach trains a low-rank update to the released student's first prediction. From initial noise, this
prediction learns to reach an intermediate state produced by the unguided base. We then disable the update and
let the unchanged student finish generation. The base supplies targets during preparation and is absent at
inference; the adapted samplers use two denoiser evaluations for FLUX.2-klein and three for SD-Turbo.
First-step teacher regression and post-hoc diversity adaptation have precedents~\cite{wu2026diversity,
liu2026crossdistill,fok2026sda}. At full ImageNet-1k scale with template prompts, adaptation improves classifier accuracy over the previously fixed base-first samplers
by \gp{inkAdVsBfKlein} points on klein and \gp{inkAdVsBfSD} on SD-Turbo (\cref{tab:main}). Matched ablations establish how target guidance, jump endpoint, and update placement affect
downstream classifier accuracy; comparisons with released models, base-first sampling, and available diversity
methods test the resulting adaptation procedure.

We evaluate three generator families across datasets, learners, and prompt regimes, and measure preparation
and generation costs. Our contributions are:
\begin{enumerate}[leftmargin=*,itemsep=1pt,topsep=2pt]
  \item \textbf{Improving released generators for visual learning.} We adapt three distilled generator
  families and compare them with the released students and base-first hybrids by training classifiers under
  the same procedure. A public SDA comparison and further adaptation of DP-DMD's released student test the approach
  alongside existing diversity methods (\cref{sec:adapter}).
  \item \textbf{Controlled evidence for the adaptation targets.} Changing guidance and replacing early
  student steps motivate the correction; matched target-guidance, endpoint, and update-placement ablations
  establish which choices improve classifier accuracy (\cref{sec:trajectory,tab:ablation}).
  \item \textbf{Scale and transfer with fixed adapters.} Adapters selected during development improve
  classifier accuracy at full ImageNet scale and transfer unchanged to fine-grained datasets. Comparisons
  across prompts, learners, and real-data availability establish the tested regimes in which adaptation helps
  (\cref{tab:main,tab:finegrained,tab:learners,tab:prompts}).
\end{enumerate}

We will release code to reproduce all experiments and to train and apply the adapter.

\section{Related Work}
\label{sec:related}

\paragraph{Synthetic data for visual learning.} Text-to-image models can supply images for training classifiers
from scratch and supplementing limited real data~\cite{sariyildiz2023fake,azizi2023synthetic,he2023synthetic}.
The resulting accuracy depends on the generator, dataset size, prompts, and learning procedure. Lower guidance
and richer descriptions improve supervised learning from synthetic images~\cite{sariyildiz2023fake,fan2024scaling},
while StableRep uses multiple images generated from the same caption as positives for self-supervised
representation learning~\cite{tian2023stablerep}. These approaches establish useful ways to generate and learn
from synthetic data. Adamkiewicz \etal~\cite{adamkiewicz2026pretty} identify a complementary problem:
images from newer and distilled generators can train weaker classifiers than images from older models.
We address this problem by adapting released distilled generators and comparing their images under matched
classifier training procedures, with additional learners testing the scope of the improvement.

\paragraph{Diffusion acceleration and distillation.} Latent diffusion and rectified-flow models support
text-conditioned generation through iterative sampling~\cite{rombach2022high,esser2024scaling}. Distillation
reduces the number of model evaluations through objectives such as adversarial training, distribution matching,
and consistency learning~\cite{sauer2024adversarial,yin2024one,yin2024improved,luo2023latent}.
Although these methods enable fast generation, maintaining visual quality does not guarantee that the images
remain equally useful for training classifiers~\cite{adamkiewicz2026pretty}. We build on their released
students, adapting a single prediction and retaining the student's unchanged completion steps. This lets us
test improvements to synthetic training data without training another distilled generator from its base.

\paragraph{Diversity preservation and adaptation.} Guidance schedules and condition annealing adjust the
trade-off between fidelity and diversity~\cite{kynkaanniemi2024applying,sadat2024cads}. Gandikota and
Bau~\cite{gandikota2026distilling} restore diversity by using base-model predictions for the initial steps of
a distilled sampler. We evaluate this base-first procedure as an alternative that requires no adapter training
but uses the base at inference. Training-based methods also distinguish early structural variation from later
refinement. DP-DMD supervises the first student step with a teacher-derived target and trains the remaining
steps through distribution matching~\cite{wu2026diversity}; CrossDistill assigns trajectory-preserving and
distribution-matching objectives to different noise intervals in video generation~\cite{liu2026crossdistill}.
CrossDistill's direct-flow variant also regresses the first prediction toward an intermediate teacher state;
our adaptation applies this supervision to a released image student while leaving its completion steps unchanged.
Other methods likewise modify distillation objectives to preserve
diversity~\cite{li2026duet,li2026distill,huang2026restoring,fan2026phased}. SDA directly adapts released few-step
models with teacher-guided directional alignment, and its Krea-2 release activates the adapter only during
early sampling steps~\cite{fok2026sda,fok2026sdazimage}.

We study how unguided teacher targets and first-prediction adaptation affect classifier accuracy when the
released student's completion steps remain fixed (\cref{tab:supp-method-positioning}).

\section{Methodology}
\label{sec:methodology}
\label{sec:adapter}

\subsection{Problem and approach}
\begin{figure*}[t]
  \centering
  \includegraphics[width=\linewidth]{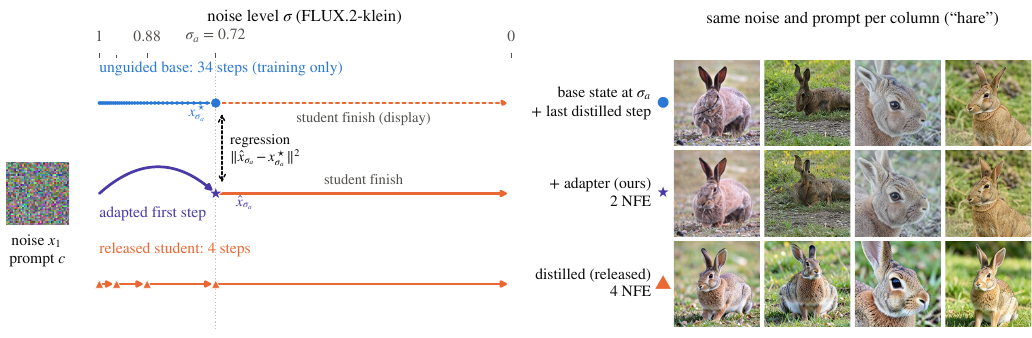}
  \caption{\textbf{One adapted prediction followed by unchanged student sampling.}
  Left: klein's released four-step student (orange), its unguided base on a fine grid (blue), and the adapted
  sampler (violet). We train the first prediction to reach the base state at $\sigma_a$; the unchanged student
  then finishes in one step, giving two denoiser evaluations without the base at inference. Right: matching
  noise and hare prompts, completed from the teacher target, by the adapter, or by the released student.
  Images are selected to illustrate the sampling paths.}
  \label{fig:method}
\end{figure*}

We adapt a released distilled generator so that classifiers trained on its images perform better on real
images. The adaptation uses the corresponding full-step base as a teacher and preserves the released student
for the remaining generation steps. We train a low-rank update to the first prediction to approximate an
intermediate teacher state, then disable the update and let the student finish. The teacher is needed to
prepare training targets but is absent at inference (\cref{fig:method}).

\subsection{Constructing the teacher target}
We compare generated datasets through the accuracy of classifiers trained on them, holding prompts, image
seeds, and classifier training fixed within each comparison. Reducing base-model guidance improves this
accuracy in the tested configurations, motivating targets that retain the base's text-conditioned prediction
without classifier-free guidance amplification. We call this setting \emph{unguided}, corresponding to a
standard guidance weight of one. The target still depends on the text prompt.

For each prompt and initial noise sample, we run the unguided base on a fine sampling grid and save its
latent state at a selected noise level. This gives paired supervision: the released student receives the same
noise and prompt and learns to predict a jump to that state. We compare guided and unguided teacher targets
in the experiments to test this choice directly (\cref{tab:ablation}).

\subsection{Correcting the early generation steps}
The target must also specify how much of the teacher trajectory the adapted prediction should replace.
Our guidance-window comparisons show that strong guidance during early steps accounts for much of its
observed cost to classifier accuracy. Replacing early student predictions with the unguided base also
improves accuracy more than replacing later predictions (\cref{sec:trajectory}). These results motivate
learning the teacher's early computation in one jump and reusing the student afterward.

We choose the jump endpoint from the released student's noise levels, allowing the unchanged student to
continue on its existing schedule. Endpoint selection uses the ImageNet-100 and 200-class development
protocols; a small ImageNet-1k pilot also contributed to method development. The selected targets and
resulting evaluation counts appear in \cref{tab:adapter-config}.
The candidate endpoints and retention criteria are specified in \cref{sec:supp-protocol}. Later evaluations
use previously fixed adapters and samplers: klein and SD-Turbo at full ImageNet scale, and all four generator
families on fine-grained datasets.

\begin{table}[t]
  \centering
  \tablebody
  \setlength{\tabcolsep}{3pt}
  \caption{\textbf{Teacher targets and adapted sampling schedules.} The first prediction jumps to the target;
  the unchanged student completes generation. NFE counts all denoiser evaluations in the adapted sampler.
  Flow-model targets use noise level $\sigma$; SD-Turbo uses its scheduler timestep $t$.}
  \label{tab:adapter-config}
\begin{tabular}{@{}llll@{}}
\toprule
Student & Base teacher & Jump endpoint & NFE \\
\midrule
FLUX.2-klein & klein base & $\sigma=0.7175$ & 2 \\
SD-Turbo & SD 2.1 & $t=499$ & 3 \\
Z-Image-Turbo & Z-Image & $\sigma=0.8571$ & 7 \\
\bottomrule
\end{tabular}

\end{table}

\subsection{Training the adapter}
Let $v_\theta(x,\sigma,c)$ denote the released student's prediction for latent $x$, noise level $\sigma$,
and prompt $c$, with sampling schedule $1=\sigma_0>\cdots>\sigma_K=0$. Starting from initial noise $x_1$,
the unguided teacher produces the target $x^\star_{\sigma_a}(x_1,c)$. We freeze the released weights and fit
low-rank updates $\Delta W=BA$, initializing $B$ to zero so that training starts from the released model.
The adapted first prediction minimizes the squared error between its resulting latent and the teacher target:
\begin{equation}
  \begin{split}
    \mathcal{L}(\Delta W)=\mathbb{E}_{x_1,c}\big\|&x_1+(\sigma_a-1)\,v_{\theta+\Delta W}(x_1,1,c)\\
    &{}-x^\star_{\sigma_a}(x_1,c)\big\|^2 .
  \end{split}
  \label{eq:adapter}
\end{equation}
This applies established first-step teacher regression~\cite{wu2026diversity,liu2026crossdistill} to a
released student. The expression uses flow-matching coordinates; SD-Turbo uses the corresponding Euler
update in its scheduler's noise coordinate (\cref{eq:sd-adapter}). Teacher-state regression trains the
adapter, while downstream classifier accuracy guides sampler development and evaluates the generated images.

We use rank-32 updates on transformer-block linear layers, or attention-block linear layers for SD-Turbo.
We prepare 16,000 noise/target pairs from ImageNet-100 class-name templates and train the primary adapters
for 3,000 optimization steps. Their noise seeds are disjoint from those used to generate evaluated
datasets. The teacher targets are prepared once and reused during optimization; optimizer settings and
family-specific sampling grids appear in \cref{sec:supp-protocol}.

\subsection{Sampling with the adapted student}
At inference, we activate the adapter for the initial prediction and jump directly to $\sigma_a$. We then
disable it and complete the remaining schedule with the released student. When the target lies beyond the
original first endpoint, this jump replaces several student steps with one prediction.

Only the student and its adapter are loaded during generation. Target preparation and adapter fitting incur
a one-time cost, which we report separately from per-image generation time and include in total deployment
cost comparisons (\cref{tab:supp-deployment}).

\section{Experiments}
\label{sec:experiments}

\subsection{Experimental setup}
\label{sec:setup}
Our primary comparison trains a ResNet-50~\cite{he2016deep} from scratch on 500 generated images per ImageNet-100 class (CMC subset~\cite{tian2020contrastive}) and
evaluates top-1 accuracy on the corresponding real validation images. Eight photo templates supply class-name
prompts, with matching prompts and image seeds across samplers. Classifiers use the same 100-epoch training
procedure within each comparison. We evaluate FLUX.2-klein, SD-Turbo, and Z-Image-Turbo against their released
students, unguided bases, and base-first samplers, alongside the public SDA adapter and DP-DMD's released
SD3.5-Medium student. Comparisons generate at 512 pixels; a separate DP-DMD check uses its documented
1024-pixel resolution with the unchanged adapter. Classifier hyperparameters, baseline settings, and the broader generator survey appear in
\cref{sec:supp-protocol}.

ImageNet-100 and the 200-class protocol provide development results, and the small ImageNet-1k pilot also
informed method development. Full-scale ImageNet-1k~\cite{deng2009imagenet} and Food-101, Cars, and Pets evaluations~\cite{bossard2014food,krause20133d,parkhi2012cats} use previously
fixed adapters and samplers: klein and SD-Turbo at full ImageNet scale, and all four generator families on
fine-grained datasets. The 200-class results include classes absent from adapter
training but were also inspected during development (\cref{tab:supp-evaluation-roles}).

Unless stated otherwise, tables and error bars report means and standard deviations over three classifier
seeds. Primary ImageNet-100 adapter results pool three adapter-training runs with three classifier seeds each;
other adapter results use one training run, with exceptions identified in their tables. Accuracy gaps
compare mean per-class accuracies, with paired 95\% class-bootstrap intervals after averaging classifier
runs, conditional on the generated datasets and fitted checkpoints. Four headline sets are regenerated
independently; the ImageNet-100 LLM comparisons use three generated datasets with three classifier seeds
each. Separate dataset-mean and classifier SDs appear in \cref{tab:supp-e50-replication}.

\subsection{Classifier accuracy from adapted images}
\begin{figure*}[t]
  \centering
  \includegraphics[width=\linewidth]{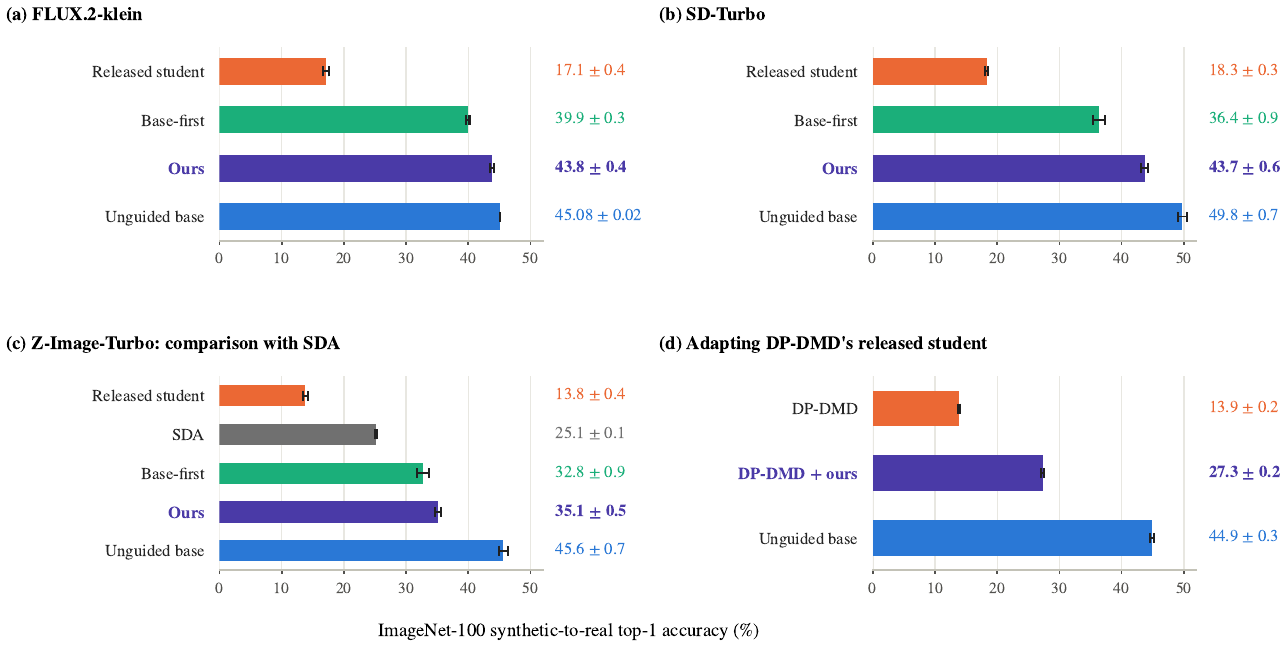}
  \caption{\textbf{First-step adaptation improves the training value of released fast generators.}
  ResNet-50 top-1 on real ImageNet-100 after training on 500 generated images/class.
  Adapters exceed released students and base-first sampling (a--c), the tested SDA adapter (c),
  and the released DP-DMD student when applied on top of it (d). Blue: unguided bases.
  Error bars: mean $\pm$ SD under the reporting convention in \cref{sec:setup}; DP-DMD plus ours uses one adapter run.
  Z-Image's released reference uses SDA's eight-step schedule; \cref{tab:supp-main-complete} also reports the nine-step student.}
  \label{fig:attribution}
\end{figure*}
On ImageNet-100, adaptation raises classifier accuracy from \tp{klein}\% to \tp{kd_ad2_x3}\% on klein and from
\tp{sdt4}\% to \tp{sdt_ad1_x3}\% on SD-Turbo. Z-Image-Turbo's adapted images reach \tp{zt_ad2_x3}\%.
\Cref{fig:attribution} compares the released students, base-first samplers, adapters, and unguided bases.
All three adapters outperform their base-first samplers, although the unguided bases retain higher accuracy
in these comparisons. \Cref{tab:main} reports the 200-class development and full ImageNet results;
complete primary accuracies and costs appear in \cref{tab:supp-main-complete}.

The SDA comparison uses its documented eight-step Z-Image-Turbo sampler. Its better tested activation
schedule reaches \tp{zt_sda_all}\%, compared with \tp{zt_ad2_x3}\% for our adapter. Further adapting DP-DMD's
released student raises accuracy from \tp{dpdmd}\% to \tp{dpdmd_ad1}\% using the selected deeper target
(\cref{fig:attribution}).
This tests whether our correction adds value to DP-DMD's student; it does not compare independently trained
distillation procedures. At its documented 1024-pixel resolution, the unchanged adapter still gains
\gp{e49native} points (\cref{tab:supp-dpdmd}). FSG, MoNO, and skipping the initial student step also give smaller improvements in
the evaluated settings (\cref{tab:ablation,sec:supp-dpdmd}).

\begin{table}[t]
  \centering\tablebody
  \setlength{\tabcolsep}{3pt}
  \caption{\textbf{The gains extend to additional classes and full ImageNet scale.}
  ResNet-50 top-1 (\%), mean $\pm$ SD over three classifier seeds, one adapter run.
  The 200-class protocol informed development (500 images/class); ImageNet-1k uses fixed samplers
  at 1,281 images/class. Bold: best fast sampler. Additional primary results, including
  ImageNet-100, COCO-target adapters, and costs, are in \cref{tab:supp-main-complete}.}
  \label{tab:main}
\begin{tabular}{@{}lrr@{}}
\toprule
Sampler & 200 classes & ImageNet-1k \\
\midrule
\multicolumn{3}{@{}l}{\emph{FLUX.2-klein}} \\
Released student & 19.1\,{\tableuncert$\pm$0.0} & 8.4\,{\tableuncert$\pm$0.1} \\
Base-first & 43.7\,{\tableuncert$\pm$0.1} & 30.8\,{\tableuncert$\pm$0.1} \\
Ours & \textbf{48.6}\,{\tableuncert$\pm$0.3} & \textbf{34.2}\,{\tableuncert$\pm$0.1} \\
Unguided base, 20 steps & 48.1\,{\tableuncert$\pm$0.2} & 37.2\,{\tableuncert$\pm$0.1} \\
\midrule
\multicolumn{3}{@{}l}{\emph{SD-Turbo}} \\
Released student & 19.8\,{\tableuncert$\pm$0.3} & 12.7\,{\tableuncert$\pm$0.2} \\
Base-first & 36.0\,{\tableuncert$\pm$0.2} & 26.7\,{\tableuncert$\pm$0.3} \\
Ours & \textbf{41.3}\,{\tableuncert$\pm$0.8} & \textbf{32.1}\,{\tableuncert$\pm$0.3} \\
Unguided base, 30 steps & 45.7\,{\tableuncert$\pm$0.9} & 44.1\,{\tableuncert$\pm$0.1} \\
\midrule
\multicolumn{3}{@{}l}{\emph{Z-Image-Turbo}} \\
Released student & 13.9\,{\tableuncert$\pm$0.4} & -- \\
Base-first & 35.3\,{\tableuncert$\pm$0.3} & -- \\
Ours & \textbf{38.2}\,{\tableuncert$\pm$0.3} & -- \\
\bottomrule
\end{tabular}

\end{table}

\Cref{tab:main} also reports full ImageNet-1k results. With 1,281 images per class, accuracy rises from \tp{in1k_klein_full}\% to
\tp{in1k_kd_ad2_full}\% on klein and from \tp{in1k_sdt4_full}\% to \tp{in1k_sdt_ad1_full}\% on SD-Turbo.
At this scale, the adapters also exceed the previously fixed base-first samplers by
\gp{inkAdVsBfKlein} points \gpci{inkAdVsBfKlein} on klein and \gp{inkAdVsBfSD} \gpci{inkAdVsBfSD} on SD-Turbo.
The gains over released and base-first sampling remain positive on the 900 classes absent from adapter training
(\cref{tab:supp-unseen-fullscale}).
Klein and SD-Turbo recover \nm{inkFullShareKlein}\% and \nm{inkFullShareSD}\% of their unguided bases' accuracy gains
over the released students, respectively.
Increasing the unadapted klein dataset from 100 to 1,281 images per class produces little change, whereas
the adapted generator benefits from the larger dataset (\cref{tab:main,sec:supp-scaling}). All dataset sizes
use the same number of classifier epochs, so larger sets receive more optimization updates.

\subsection{Generation and preparation costs}
\begin{figure*}[t]
  \centering
  \includegraphics[width=\linewidth]{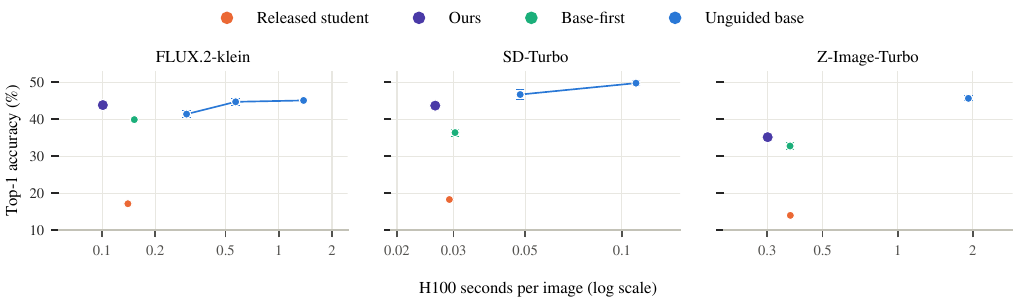}
  \caption{\textbf{Adaptation improves classifier accuracy while retaining fast generation.}
  ImageNet-100 top-1 against measured H100 seconds/image (log scale); vertical bars show SD.
  Unguided-base points use 10, 20, and 50 steps for klein, 10 and 30 for SD~2.1, and 50 for Z-Image.
  The adapted samplers require only the student at inference. Preparation is excluded here and included
  in \cref{tab:supp-deployment}; additional operating points appear in \cref{fig:supp-cost-full}.}
  \label{fig:cost}
\end{figure*}
We measure generation throughput on one H100 at the recorded batch sizes, including prompt encoding,
denoising, and decoding. These measurements exclude model loading and writing generated images to storage.
The adapted klein sampler takes \tpsec{kd_ad2_x3} seconds per image, compared with
\tpsec{kb_g1_s20} for its unguided 20-step base. Across the three primary families, adapted sampling costs
match or improve on the released students while increasing classifier accuracy (\cref{fig:cost}).

Preparing teacher targets and fitting the first klein and SD-Turbo adapters costs \nm{prepHoursKlein} and
\nm{prepHoursSD} H100-hours, respectively. At 50,000 generated images, preparation makes adaptation more
expensive overall than base-first, while the adapter gives higher accuracy and requires no base at inference.
With template-prompt throughputs measured on ImageNet-100, adaptation becomes cheaper overall than base-first after approximately
\nm{breakEvenBaseFirstKlein} images for klein and \nm{breakEvenBaseFirstSD} for SD-Turbo (\cref{tab:supp-deployment}). We exclude SDA runtime from
the speed comparison because our unmerged implementation does not represent an optimized deployment.

\subsection{Evidence for the design choices}
\label{sec:trajectory}
\begin{table}[t]
  \centering
  \tablebody
  \setlength{\tabcolsep}{4pt}
  \renewcommand{\arraystretch}{1.02}
  \caption{\textbf{Target choice matters more than parameter count.}
ImageNet-100 ResNet-50 top-1 (\%). Ours and SD-Turbo rank 128 pool
three adapter runs; others use one. Klein's guided-target and endpoint variants use the finer teacher grid
(second row); the deeper endpoint is 0.511. Bold: best per column. FSG, MoNO (published basis: four images), and skipping step 1 require
no training. ``Adapter in every step'' reuses the first-step-trained checkpoint throughout the adapted
sampling schedule: two evaluations for klein and three for SD-Turbo.}
  \label{tab:ablation}
\begin{tabular}{@{}lrr@{}}
\toprule
variant & FLUX.2-klein & SD-Turbo \\
\midrule
ours: unguided, step 1, rank 32 & 43.8\,{\tableuncert$\pm$0.4} & 43.7\,{\tableuncert$\pm$0.6} \\
\quad same, finer teacher grid & 43.2\,{\tableuncert$\pm$0.7} & -- \\
\addlinespace[2pt]
target at guidance 2 & 31.5\,{\tableuncert$\pm$0.7} & 41.2\,{\tableuncert$\pm$0.9} \\
target at default guidance & 20.1\,{\tableuncert$\pm$0.4} & 30.4\,{\tableuncert$\pm$0.4} \\
\addlinespace[2pt]
adapter in every step & 41.4\,{\tableuncert$\pm$0.3} & 38.8\,{\tableuncert$\pm$0.8} \\
every weight fine-tuned & 43.2\,{\tableuncert$\pm$1.0} & 43.4\,{\tableuncert$\pm$0.3} \\
rank 8 & 43.4\,{\tableuncert$\pm$0.3} & 43.3\,{\tableuncert$\pm$0.3} \\
rank 128 & \textbf{44.1}\,{\tableuncert$\pm$0.4} & \textbf{44.4}\,{\tableuncert$\pm$0.6} \\
\addlinespace[2pt]
jump to the first level only & 32.1\,{\tableuncert$\pm$0.6} & 32.2\,{\tableuncert$\pm$0.4} \\
jump to a deeper level & 40.0\,{\tableuncert$\pm$0.5} & 41.2\,{\tableuncert$\pm$0.7} \\
\addlinespace[2pt]
FSG & 20.7\,{\tableuncert$\pm$0.7} & -- \\
MoNO, 4 images per basis & 26.8\,{\tableuncert$\pm$0.3} & -- \\
MoNO, 32 images per basis & 29.9\,{\tableuncert$\pm$0.3} & -- \\
\addlinespace[2pt]
skip first step (no training) & 31.8\,{\tableuncert$\pm$0.3} & 14.4\,{\tableuncert$\pm$0.7} \\
distilled model alone & 17.1\,{\tableuncert$\pm$0.4} & 18.3\,{\tableuncert$\pm$0.3} \\
\bottomrule
\end{tabular}

\end{table}
Changing guidance only during early or late generation steps shows that the early part accounts for much
of the observed accuracy loss. Likewise, replacing early student steps with the unguided base improves
accuracy more than replacing later steps. Detailed guidance-window and base/student comparisons support
the early-step design (\cref{fig:trajectory}).

The adapter ablations then compare the teacher targets directly (\cref{tab:ablation}). With the same finer
teacher grid on klein, an unguided target gives \tp{kd_ad2x}\%, while a guidance-four target gives
\tp{kd_ad2_tg4}\%. SD-Turbo and Z-Image show the same direction of change (\cref{sec:supp-additional}). Stopping the adapted jump at the
original first endpoint also underperforms the selected target on klein and SD-Turbo, as do the tested
deeper endpoints. Keeping the same first-step-trained adapter active during the remaining sampling steps
lowers accuracy. This comparison tests inference-time activation; it does not evaluate an adapter trained
across timesteps. Increasing rank or training every weight gives comparatively small changes.
Together, these comparisons support an unguided
intermediate target and an update used only for the initial prediction.

\subsection{Generalization across datasets and learners}
\begin{table}[t]
  \centering
  \tablebody
  \setlength{\tabcolsep}{3pt}
  \renewcommand{\arraystretch}{1.02}
  \caption{\textbf{Adapters transfer unchanged to fine-grained datasets.}
ResNet-50 top-1 (\%), 500 generated images/class; mean $\pm$ SD over three classifier seeds. Our adapters use one training run.
NFE: denoiser evaluations. Bold: best shown per family. Additional references, gaps, and runtimes:
\cref{tab:supp-finegrained-references,tab:supp-crossdata}.}
  \label{tab:finegrained}
\begin{tabular}{@{}lrrrr@{}}
\toprule
Sampler & NFE & Food-101 & Cars & Pets \\
\midrule
\multicolumn{5}{@{}l}{\emph{FLUX.2-klein}} \\
Released & 4 & 10.5\,{\tableuncert$\pm$0.3} & 18.9\,{\tableuncert$\pm$0.4} & 20.5\,{\tableuncert$\pm$1.0} \\
Base-first & 4 & 25.9\,{\tableuncert$\pm$0.1} & \textbf{38.8}\,{\tableuncert$\pm$0.3} & 45.3\,{\tableuncert$\pm$0.5} \\
Ours & 2 & \textbf{30.9}\,{\tableuncert$\pm$0.2} & 36.2\,{\tableuncert$\pm$0.2} & \textbf{46.2}\,{\tableuncert$\pm$0.8} \\
\midrule
\multicolumn{5}{@{}l}{\emph{SD-Turbo}} \\
Released & 4 & 10.0\,{\tableuncert$\pm$0.3} & 46.9\,{\tableuncert$\pm$0.4} & 19.5\,{\tableuncert$\pm$1.4} \\
Base-first & 4 & 27.3\,{\tableuncert$\pm$0.2} & 64.8\,{\tableuncert$\pm$0.5} & 50.6\,{\tableuncert$\pm$0.9} \\
Ours & 3 & \textbf{31.8}\,{\tableuncert$\pm$0.2} & \textbf{68.3}\,{\tableuncert$\pm$0.2} & \textbf{54.5}\,{\tableuncert$\pm$0.9} \\
\midrule
\multicolumn{5}{@{}l}{\emph{Z-Image-Turbo}} \\
Released & 9 & 9.2\,{\tableuncert$\pm$0.3} & 27.6\,{\tableuncert$\pm$0.3} & 15.4\,{\tableuncert$\pm$0.6} \\
SDA & 8 & 16.8\,{\tableuncert$\pm$0.3} & 54.6\,{\tableuncert$\pm$0.2} & 37.8\,{\tableuncert$\pm$0.4} \\
Ours & 7 & \textbf{23.4}\,{\tableuncert$\pm$0.5} & \textbf{60.0}\,{\tableuncert$\pm$0.1} & \textbf{49.4}\,{\tableuncert$\pm$0.9} \\
\midrule
\multicolumn{5}{@{}l}{\emph{DP-DMD's student}} \\
Released & 4 & 6.9\,{\tableuncert$\pm$0.2} & 45.4\,{\tableuncert$\pm$0.7} & 12.5\,{\tableuncert$\pm$0.3} \\
+ ours & 3 & \textbf{21.8}\,{\tableuncert$\pm$0.4} & \textbf{56.4}\,{\tableuncert$\pm$0.7} & \textbf{46.6}\,{\tableuncert$\pm$0.8} \\
\bottomrule
\end{tabular}

\end{table}
The ImageNet-trained adapters improve classifiers on Food-101, Cars, and Pets without retraining or changing
their sampling settings (\cref{tab:finegrained}). Our Z-Image-Turbo adapter exceeds SDA on all three datasets,
and our adapter also improves DP-DMD's released student on each dataset. The class-bootstrap intervals for
all these gains lie above zero (\cref{tab:supp-crossdata}). SD-Turbo's adapter exceeds base-first throughout;
klein's trails base-first on Cars. COCO captions can also replace class-name adaptation prompts
(\cref{tab:supp-main-complete,tab:supp-finegrained-references}).

\begin{table}[t]
  \centering
  \tablebody
  \setlength{\tabcolsep}{2pt}
  \renewcommand{\arraystretch}{1.02}
  \caption{\textbf{Gains persist across learner architectures and recipes.}
ImageNet-100 top-1 (\%), mean $\pm$ sd over three seeds, one adapter run. Learners train from scratch on the same
sets. R50: ResNet-50; RA: RandAugment; prior recipe:~\cite{adamkiewicz2026pretty}.
Bold: adapted versus released generator. Full references: \cref{tab:supp-learners}.}
  \label{tab:learners}
\begin{tabular}{@{}lr@{\hspace{7pt}}rr@{\hspace{7pt}}r@{}}
\toprule
learner & \multicolumn{2}{c}{FLUX.2-klein} & \multicolumn{2}{c}{SD-Turbo} \\
\cmidrule(lr){2-3}\cmidrule(l){4-5}
 & Original & Adapted & Original & Adapted \\
\midrule
R50 + RA & 22.0\,{\tableuncert$\pm$0.3} & \textbf{46.7}\,{\tableuncert$\pm$0.4} & 23.4\,{\tableuncert$\pm$0.7} & \textbf{48.0}\,{\tableuncert$\pm$0.6} \\
Prior recipe & 17.9\,{\tableuncert$\pm$1.0} & \textbf{39.0}\,{\tableuncert$\pm$0.3} & 21.7\,{\tableuncert$\pm$1.2} & \textbf{38.6}\,{\tableuncert$\pm$0.5} \\
ConvNeXt-T & 19.9\,{\tableuncert$\pm$0.2} & \textbf{39.0}\,{\tableuncert$\pm$0.2} & 20.7\,{\tableuncert$\pm$0.5} & \textbf{41.8}\,{\tableuncert$\pm$1.4} \\
ViT-S/16 & 15.6\,{\tableuncert$\pm$0.1} & \textbf{26.9}\,{\tableuncert$\pm$0.5} & 13.8\,{\tableuncert$\pm$0.4} & \textbf{26.8}\,{\tableuncert$\pm$0.3} \\
\bottomrule
\end{tabular}

\end{table}

The gains persist with RandAugment, the reimplemented classifier recipe of~\cite{adamkiewicz2026pretty},
ConvNeXt-T, and ViT-S/16 (\cref{tab:learners}). Each comparison trains released-image and adapted-image
classifiers under the same procedure. Their different gain sizes show that the learning recipe affects the
magnitude, while the positive results support improvements beyond the primary ResNet-50 setup. Features
learned from adapted images also transfer better to additional recognition tasks, and full-scale gains
persist on shifted ImageNet test sets (\cref{tab:transfer,tab:supp-ood}).

\subsection{Prompt regimes and real-data availability}
\begin{table}[t]
  \centering
  \tablebody
  \setlength{\tabcolsep}{3pt}
  \caption{\textbf{The benefit depends on fitting prompts and evaluation regime.}
  Top: ImageNet-100 gains (points [95\% class-bootstrap CI]), generation/fitting prompts, 500 images/class.
  LLM rows pool three datasets; others use one, each with three classifier seeds. Bottom: ImageNet-1k,
  100 images/class, one dataset and three classifier seeds, versus released students using LLM prompts.
  Template-based: template fitting/generation. LLM-prompt: LLM generation, fitting selected on ImageNet-100
  (LLM for klein, templates for SD-Turbo). One training run per adapter. Bold: interval above zero.}
  \label{tab:prompts}
\begin{tabular}{@{}lrr@{}}
\toprule
Comparison & FLUX.2-klein & SD-Turbo \\
\midrule
\multicolumn{3}{@{}l}{\emph{ImageNet-100: adapted $-$ released}} \\
Templates / templates & \textbf{+26.9}\,{\tableuncert[+23.5, +30.4]} & \textbf{+25.0}\,{\tableuncert[+21.9, +28.1]} \\
Captions / templates & \textbf{+6.4}\,{\tableuncert[+4.3, +8.6]} & \textbf{+20.7}\,{\tableuncert[+17.7, +23.8]} \\
LLM / templates & \textbf{+2.1}\,{\tableuncert[+0.2, +4.0]} & \textbf{+16.7}\,{\tableuncert[+14.1, +19.4]} \\
LLM / LLM & \textbf{+5.4}\,{\tableuncert[+3.4, +7.3]} & \textbf{+15.6}\,{\tableuncert[+12.9, +18.5]} \\
\midrule
\multicolumn{3}{@{}l}{\emph{ImageNet-1k: vs. released with LLM prompts}} \\
\shortstack[l]{Template-based\\procedure} & \textbf{+2.7}\,{\tableuncert[+2.0, +3.4]} & \textbf{+7.1}\,{\tableuncert[+6.3, +7.9]} \\
\shortstack[l]{LLM-prompt\\procedure} & $-$1.5\,{\tableuncert[$-$2.0, $-$1.0]} & \textbf{+6.9}\,{\tableuncert[+6.2, +7.7]} \\
\bottomrule
\end{tabular}

\end{table}
Our template-based procedure remains useful against richer-prompt alternatives. At ImageNet-1k with
100 images per class, it exceeds released students using image-free LLM descriptions by
\nm{e50OriginalKleinVsLlmReleased} points for klein and \nm{e50OriginalSDVsLlmReleased} for SD-Turbo,
and also exceeds base-first with LLM prompts (\cref{tab:prompts,tab:supp-e50-in1k}). These compare complete
procedures with different generation prompts. Real-image captions also retain gains but require those
images (\cref{tab:supp-captions}).

The LLM-fitted adapters give higher ImageNet-100 accuracy for klein and lower accuracy for SD-Turbo than
their template-fitted counterparts; both perform worse with template generation (\cref{tab:supp-e50-fitting}).
Under matched LLM prompts on ImageNet-1k, SD-Turbo's template-trained adapter exceeds released and
base-first references, whereas klein's selected LLM-fitted adapter and post-hoc template-trained variant
fall below both. These adapter comparisons retain their directions on the 900 classes outside fitting
(\cref{tab:supp-e50-in1k}). Klein's unguided 20-step base also falls \nm{e52KleinBaseLoss} points below its
released student, whereas unguided SD~2.1 exceeds SD-Turbo by \nm{e52SDBaseGain} points (\cref{tab:supp-e50-in1k}).

With limited real training data, adding adapted images improves real-only accuracy in the tested mixtures.
With the full real ImageNet-1k training set, adding adapted images improves Sketch and R accuracy while
slightly lowering in-distribution accuracy. The full results in the supplement delimit the settings in which
synthetic augmentation helps (\cref{tab:supp-ood,sec:supp-mixtures}).

\section{Discussion}
\label{sec:discussion}

\paragraph{The teacher target shapes downstream learning.}
Learning a jump to the selected unguided state gives higher classifier accuracy than matching a guided
target or retaining the original first endpoint (\cref{tab:ablation}). Further adapting DP-DMD's released
student also improves classifier accuracy, although a substantial gap to its unguided base remains
(\cref{sec:supp-dpdmd}). The evaluated comparisons do not isolate the source of that remaining gap.

\paragraph{The useful choice depends on the generator and prompts.}
Template-based adaptation improves on released and base-first samplers on ImageNet-100 and on their
LLM-prompt alternatives at ImageNet-1k with 100 images per class (\cref{fig:attribution,tab:supp-e50-in1k}).
For klein at this dataset size, we therefore recommend the original adapter with template generation;
SD-Turbo benefits from adaptation with either prompt source. Released klein with LLM prompts avoids adapter
preparation, and base-first improves its accuracy when the base can be loaded. Captions derived from real images
retain gains for both models but require access to those images (\cref{tab:supp-captions}). Dataset also
matters: base-first exceeds klein's adapter on Cars (\cref{tab:finegrained}).

\paragraph{Fast generation and low total cost are separate benefits.}
Once fitted, the adapter improves classifier accuracy at comparable or lower generation time than the
released student, without the base at inference (\cref{fig:cost}). Preparation adds an initial cost: at
50,000 images, base-first is cheaper overall in the measured klein and SD-Turbo template settings, while
adaptation gives higher accuracy (\cref{tab:supp-deployment}). At full ImageNet scale, estimated preparation plus
generation costs \nm{fullTotalAdKlein} H100-hours for klein's adapter versus \nm{fullTotalBfKlein} for base-first.
SD-Turbo's costs are approximately equal, while adaptation gives higher classifier accuracy in both families
(\cref{tab:main,tab:supp-deployment}). Adaptation is useful when this gain justifies
preparation, the adapter can be reused, or inference requires only the student. The unguided base remains
an option when its higher accuracy justifies slower sampling.

\paragraph{Set metrics help diagnose generators.}
Real-data recall and CLIP Fr\'echet distance track classifier accuracy across the studied configurations.
Family-weighting and held-out-family analyses support these set-level diagnostics, with varying predictive
performance across families (\cref{tab:supp-metrics,tab:supp-family-weighting,tab:supp-heldout}). Shared prompts,
seeds, and neighboring settings limit independence. The associations establish neither causality nor reliable
classwise prediction; training classifiers remains the direct evaluation of usefulness.

\section{Limitations \& Future Work}
\label{sec:limitations}

Our method requires a compatible base model to construct targets and an intermediate state that the released
student can complete. We select the endpoint empirically from the student's schedule using downstream
classifier results. We do not establish a rule that selects a useful endpoint for a new generator without
this evaluation. Developing such a rule, or learning
targets that better match the frozen student continuation, could reduce the cost of applying the method to
additional generators.

The evaluation also has a development boundary. ImageNet-100, the 200-class protocol, and the small
ImageNet-1k pilot informed method choices; they are not independent final tests. Full-scale ImageNet-1k and
fine-grained evaluations use previously fixed adapters and samplers, but classes absent from adapter fitting
can still have appeared during development (\cref{tab:supp-evaluation-roles}). Most reported variation concerns
classifier training, while adapter-training variation and independent dataset regeneration are measured for
subsets of the comparisons. Repeating more comparisons with independently generated datasets would better
characterize their uncertainty.

We evaluate classification and feature transfer; detection and segmentation remain untested. Prompt-specific fitting is evaluated with one training run per adapter. The rich-prompt
ImageNet-1k evaluation uses 100 images per class and one generated dataset per sampler, so it does not
establish the outcome at the full synthetic dataset size. Understanding how prompt distributions interact
with the adapted prediction and frozen completion steps could improve transfer across prompt regimes.
Additional tasks require accuracy and preparation-cost evaluation. Adaptation also inherits the biases and
coverage limitations of the released generators. Aggregate classification gains do not establish balanced
performance across demographic groups or rare visual conditions, which require separate evaluation.

\section{Conclusion}
\label{sec:conclusion}

We adapt released distilled generators to produce images that train more accurate visual classifiers.
A low-rank update teaches the initial prediction to reach an unguided intermediate base-model state;
the unchanged student then completes generation without the base at inference. Controlled comparisons
establish the importance of target guidance, endpoint, and update placement. The improvements extend across
three generator families, with fixed klein and SD-Turbo adapters also benefiting full-scale ImageNet training
and fine-grained recognition. These results provide a simple way to improve training data from released
generators while retaining fast generation. Prompt choice, preparation cost, and inference requirements
determine the most useful configuration.

\label{maintextend}

\clearpage
\clearpage
\onecolumn
\appendix
\setcounter{table}{0}
\setcounter{figure}{0}
\renewcommand{\thetable}{S\arabic{table}}
\renewcommand{\thefigure}{S\arabic{figure}}
\renewcommand{\floatpagefraction}{0.6}
\renewcommand{\dblfloatpagefraction}{0.6}
\raggedbottom
\begin{center}\Large\bfseries Supplementary material\end{center}
\noindent This supplement follows the main paper: protocols, target design, baseline comparisons, generalization,
prompt and real-data regimes, and supporting diagnostics. Figures report the additional guidance-window, handoff, baseline, and data-scaling experiments.
Tables provide numerical results, reference accuracies, and diagnostic sensitivity checks; the main
comparison plots are not reproduced.

\FloatBarrier\section{Reproduction and evaluation protocols}
\label{sec:supp-protocol}
This section specifies dataset roles, classifier training, generator settings, adapter fitting, and cost accounting.

\par\medskip\noindent
\begin{minipage}{\linewidth}
  \captionsetup{type=table,hypcap=false}
  \centering
  \tablebody
  \setlength{\tabcolsep}{4pt}
  \renewcommand{\arraystretch}{1.08}
  \begin{tabular}{@{}p{0.22\textwidth}p{0.32\textwidth}p{0.42\textwidth}@{}}
  \toprule
  \raggedright Protocol & \raggedright Role & \raggedright Influence on method choices \tabularnewline
  \midrule
  \raggedright ImageNet-100 & \raggedright Development and reported performance & \raggedright Shared validation images select sampler settings, endpoints, and update design. \tabularnewline
  \raggedright 200-class protocol & \raggedright Cross-class development evidence & \raggedright Multiple endpoints inspected for klein, SD-Turbo, and Z-Image; supported retaining SD-Turbo timestep 499 over 249. \tabularnewline
  \raggedright ImageNet-1k\newline100 images/class, templates & \raggedright Scale pilot & \raggedright Contributed to initial method validation. \tabularnewline
  \raggedright ImageNet-1k\newline100 images/class, LLM prompts & \raggedright Fixed-setting prompt evaluation & \raggedright Adapter choice follows the pre-specified ImageNet-100 rule and is frozen before ImageNet-1k classifier training. Classes also appeared in the template-prompt pilot. \tabularnewline
  \raggedright ImageNet-1k\newline1,281 images/class & \raggedright Fixed-setting scale evaluation & \raggedright Previously fixed klein/SD-Turbo adapters and samplers; no changes from these results. The 900-class subgroup excludes adapter-training classes. \tabularnewline
  \raggedright Food-101, Cars, Pets & \raggedright Fixed-setting cross-dataset evaluation & \raggedright Fixed adapters and samplers for all four generator families; no endpoint, rank, or weight changes from these results. \tabularnewline
  \bottomrule
  \end{tabular}
  \caption{\textbf{Later fixed-setting evaluations test choices made during development.}
  Roles in the evaluation sequence. Classes unseen in adapter training can still participate in method development,
  as in the 200-class protocol. Fixed-setting results evaluate previously selected adapters and samplers;
  the full-scale 900-class subgroup retains the original 1,000-way predictions (\cref{tab:supp-unseen-fullscale}).}
  \label{tab:supp-evaluation-roles}
\end{minipage}\par\medskip

\subsection{Datasets, classifiers, and repetition}
\paragraph{Classifier training.} The default learner is a ResNet-50~\cite{he2016deep} trained from scratch for
100 epochs. We use SGD with Nesterov momentum 0.9, learning rate 0.2 at batch size 256, five warmup epochs followed
by cosine decay, weight decay $10^{-4}$, label smoothing 0.1, and bf16 autocast. Training uses random resized crops
to 224 pixels and horizontal flips; evaluation resizes to 256 and center-crops to 224. The ViT-S/16 uses AdamW with
learning rate $5\cdot10^{-4}$, weight decay 0.05, ten warmup epochs, and RandAugment (2, 9).

\paragraph{Alternative classifier recipe.} The recipe of~\cite{adamkiewicz2026pretty}, used in
\cref{tab:learners,tab:supp-learners}, uses Adam at $10^{-3}$, batch size 1,024, 80 epochs, CutMix, gradient
clipping, and the published augmentations. It selects the checkpoint with lowest loss on a held-out tenth of the
generated images. Our primary comparisons use the fixed 100-epoch schedule above.

\paragraph{Repetitions and reporting.} Unless stated otherwise, each generated set is evaluated with three classifier seeds, and tables report their mean and standard deviation. ImageNet-100 results for the three primary adapters and the first-level DP-DMD adapter pool three adapter-training seeds with three classifier seeds each. Transfer evaluations use the first trained adapter. Ablations use one adapter-training seed, except SD-Turbo rank 128, which uses three. Data-scaling evaluations use two classifier seeds, with the exception identified in the scaling caption. We assess generation variability by independently regenerating four headline sets. Top-1 averages test images. Gaps use mean per-class accuracy and paired class-bootstrap intervals after averaging classifier runs. Mean-class and image-averaged accuracy can differ on imbalanced test splits. The intervals condition on the observed datasets and fitted adapters; classifier and adapter runs do not count as independent dataset regenerations.

\paragraph{Evaluation splits and real-data references.} ImageNet classifiers are evaluated on the real validation
images of their classes, 50 per class. The default real-data reference draws 500 images per class from the training
split. Fine-grained synthetic sets also contain 500 images per class, but evaluation uses each dataset's test split.
Their real-data references use 500 real images per class for Food-101, all 8,144 training images for Cars, and all
3,680 for Pets. These references train for the same number of optimization steps as the generated sets.

For representation transfer, we freeze the synthetic-trained ResNet-50, standardize its features, and select the
linear probe's L2 penalty on a held-out tenth of the target training split. Target images train only the probe.

\subsection{Sampler development and adapter selection}
\paragraph{Development data.} Sampler selection and reported ImageNet-100 accuracy use the same
5,000 real validation images. \Cref{tab:supp-evaluation-roles} distinguishes the development protocols from the
later fixed-setting evaluations. In particular, the 200-class protocol and the ImageNet-1k scale pilot informed
method development; classes absent from adapter fitting are not necessarily untouched during development.

\paragraph{Endpoints and rank.}
For klein, we evaluate target noise levels 0.958, 0.884, 0.7175, and 0.511; for SD-Turbo, timesteps 749, 499,
and 249. Subsequent endpoint ablations retain the klein and SD-Turbo choices: no alternative clears the specified
replacement criterion of more than 1.5 ImageNet-100 points at no more denoiser evaluations. The conditional
200-class check for such a replacement was therefore not triggered. Rank 32 is retained when within one point
of the best rank-8, rank-128, or full-update alternative.
For Z-Image, the initial comparison selects 0.857 over 0.913; later targets 0.7895 and 0.7059 raise accuracy by
less than one point and reduce its margin over the corresponding base-first sampler, so the reported endpoint
is retained.
Base guidance and hand-off settings use the tested candidate with highest ImageNet-100 top-1 within the stated
role; their candidate sets are specified in the machine-readable manifest.

Adapter optimization uses a disjoint 5\% of noise/target pairs (at least 64) to measure state-regression error,
with a fixed split seed of zero. This held-out error selects between the two full-update learning rates;
standard low-rank adapters use the final 3,000-step checkpoint. This pair split is distinct from the downstream
real-image validation set used to develop the sampler. The alternative classifier recipe selects its checkpoint
on held-out generated images; our default classifier uses the final fixed-epoch model.

\subsection{Feature metrics and cost measurement}
\paragraph{Feature metrics.} DINOv3 features use the ViT-L/16 pretrained on LVD-1689M: the 1,024-dimensional final
normalized class token, after bilinear resizing of the short side to 256 and center-cropping to 224 with ImageNet channel normalization.
CLIP uses OpenAI ViT-L/14 projected image embeddings (768 dimensions), bicubic resize/crop to 224 and CLIP channel
normalization. We L2-normalize both feature types. Precision, recall, density, and coverage use Euclidean distances
within each class, with the fifth-nearest-neighbor radius excluding self-distances ($k=5$); metrics average over
100 classes with 500 real training and 500 generated images each. Recall is the fraction of real features inside
the union of generated-feature balls. Fr\'echet distances use whole-set means and unbiased covariances. FID uses
2,048-dimensional Inception-v3 pool3 features with TensorFlow-FID weights, bicubic resize/crop to 299, and the same
50,000 real training images as reference. Preference metrics use 5,000 evenly spaced images per set: the LAION
improved-aesthetic head on normalized CLIP ViT-L/14 embeddings and PickScore v1 scores for image and prompt pairs.

\paragraph{Timing and amortization.} Timings use one H100. The klein and Z-Image students, adapters, and
base-first hybrids use batches of eight; SD-Turbo uses 32, while its unguided SD~2.1 base uses 16. Klein/Z-Image
run in bfloat16 and SD in float16; the low-rank update is evaluated in float32. Each recorded batch interval starts
immediately before the text-to-image pipeline call and ends after CUDA synchronization. It includes prompt
encoding, denoising, and VAE decoding, and excludes model loading, image encoding, and writing packs. We average
per-image times over the generated images, including the first batch; no separate warm-up batches are discarded.
Legacy default-model timings use the first 256 images at the recorded settings. These are measured pipeline
throughputs at the stated batch sizes, not batch-one latency or separately optimized hardware limits.
FSG timings include its feature optimization. MoNO generates images sequentially and includes noise optimization,
generation, and feature-history updates, with prompt encoding amortized over its input batch. Exact batch sizes
for the other generators are in the manifest.

The cost for $M$ generated images is $C(M)=C_{\rm pairs}+C_{\rm fit}+M\,t_{\rm image}$, counting preparation once per
released adapter. Klein's \nm{prepHoursKlein} H100-hours of preparation amortize against its unguided 20-step base
at approximately \nm{breakEvenKlein} images; SD-Turbo's \nm{prepHoursSD} hours amortize against unguided SD~2.1
at approximately \nm{breakEvenSD} images. Against base-first, the corresponding crossovers are approximately
\nm{breakEvenBaseFirstKlein} images for klein and \nm{breakEvenBaseFirstSD} for SD-Turbo under template prompts,
using ImageNet-100 throughputs. Full-scale ImageNet-1k throughputs instead give approximately
\nm{e51CrossoverMillionKlein} million and \nm{e51CrossoverMillionSD} million images, respectively.
SD-Turbo's estimate is sensitive to throughput variation because the adapter and base-first times differ little.
These estimates use recorded per-image times and
job-level preparation costs, assuming constant throughput and counting preparation once. They exclude sampler
development and downstream classifier evaluation. They compare total generation costs, not equal-accuracy
operating points; adapting the student improves accuracy over its released version without requiring the base
at inference.

\par\medskip\noindent
\begin{minipage}{\linewidth}
  \captionsetup{type=table,hypcap=false}
  \centering
  \tablebody
  \setlength{\tabcolsep}{5pt}
\begin{tabular}{@{}lrrrrrl@{}}
\toprule
Sampler & \multicolumn{2}{c}{One-time preparation (h)} & Generation & Total & Top-1 & Base at \\
\cmidrule(lr){2-3}
 & Targets & Fit & s/image & H100-hours & (\%) & inference \\
\midrule
\multicolumn{7}{@{}l}{\emph{FLUX.2-klein, ImageNet-100 (total for 50,000 images)}} \\
Released student & -- & -- & 0.140 & 1.9 & 17.1\,{\tableuncert$\pm$0.4} & No \\
Base-first & -- & -- & 0.152 & 2.1 & 39.9\,{\tableuncert$\pm$0.3} & Yes \\
+ adapter (first run) & 4.6 & 1.1 & 0.101 & 7.1 & \textbf{44.1}\,{\tableuncert$\pm$0.5} & No \\
Unguided base, 20 steps & -- & -- & 0.570 & 7.9 & 44.7\,{\tableuncert$\pm$0.9} & Yes \\
\midrule
\multicolumn{7}{@{}l}{\emph{SD-Turbo / SD 2.1, ImageNet-100 (total for 50,000 images)}} \\
Released student & -- & -- & 0.029 & 0.4 & 18.3\,{\tableuncert$\pm$0.3} & No \\
Base-first & -- & -- & 0.030 & 0.4 & 36.4\,{\tableuncert$\pm$0.9} & Yes \\
+ adapter (first run) & 0.7 & 0.2 & 0.026 & 1.3 & \textbf{43.3}\,{\tableuncert$\pm$0.2} & No \\
Unguided base, 30 steps & -- & -- & 0.111 & 1.5 & 49.8\,{\tableuncert$\pm$0.7} & Yes \\
\midrule
\multicolumn{7}{@{}l}{\emph{FLUX.2-klein, ImageNet-1k (total for 1,281,000 images)}} \\
Released student & -- & -- & 0.139 & 49.6 & 8.4\,{\tableuncert$\pm$0.1} & No \\
Base-first & -- & -- & 0.153 & 54.5 & 30.8\,{\tableuncert$\pm$0.1} & Yes \\
+ adapter (first run) & 4.6 & 1.1 & 0.101 & 41.4 & \textbf{34.2}\,{\tableuncert$\pm$0.1} & No \\
Unguided base, 20 steps & -- & -- & 0.571 & 203.3 & 37.2\,{\tableuncert$\pm$0.1} & Yes \\
\midrule
\multicolumn{7}{@{}l}{\emph{SD-Turbo, ImageNet-1k (total for 1,281,000 images)}} \\
Released student & -- & -- & 0.029 & 10.3 & 12.7\,{\tableuncert$\pm$0.2} & No \\
Base-first & -- & -- & 0.029 & 10.4 & 26.7\,{\tableuncert$\pm$0.3} & Yes \\
+ adapter (first run) & 0.7 & 0.2 & 0.026 & 10.2 & \textbf{32.1}\,{\tableuncert$\pm$0.3} & No \\
\bottomrule
\end{tabular}

  \caption{\textbf{Preparation, generation cost, accuracy, and model requirements define the operating point.}
  Additional preparation and generation on one H100, with ResNet-50 top-1 on each panel's protocol (mean $\pm$ SD over three
  classifier seeds; the first adapter run). Targets/fit: one-time target construction/optimization in H100-hours;
  --: no adapter preparation. Total: approximate preparation plus generation for the panel's image count, excluding classifier
  training, model loading during generation, and image writing, at the recorded batch sizes. ImageNet-1k panels use
  throughputs measured on their full-scale sets. Bold: adapter versus released student. These are observed operating points,
  not equal-accuracy cost comparisons.}
  \label{tab:supp-deployment}
\end{minipage}\par\medskip

\paragraph{Cost across dataset sizes.} \Cref{tab:supp-e50-costs} reports preparation plus generation for template
and LLM prompts at three dataset sizes. Fitting on LLM descriptions also requires generating the adaptation
description pool; its cost is charged in full to each adapter even though both share it. Construction of
generation-prompt pools is excluded for every sampler. The unguided-base
LLM estimates reuse template-prompt throughput because those reference samplers were not timed with LLM prompts.

\par\medskip\noindent
\begin{minipage}{\linewidth}
  \captionsetup{type=table,hypcap=false}
  \centering\footnotesize
  \setlength{\tabcolsep}{3pt}
\begin{tabular}{@{}llrrrrrr@{}}
\toprule
Sampler & Prompts & Prep. h & 50k & 500k & 1.28M & $N^*_{\rm BF}$ & $N^*_{\rm base}$ \\
\midrule
\multicolumn{8}{@{}l}{\emph{FLUX.2-klein}} \\
released & Templates & 0.00 & 1.9 & 19.4 & 49.8 & -- & -- \\
released & LLM & 0.00 & 1.9 & 19.4 & 49.8 & -- & -- \\
base-first & Templates & 0.00 & 2.1 & 21.1 & 54.2 & -- & -- \\
base-first & LLM & 0.00 & 2.1 & 21.2 & 54.3 & -- & -- \\
template adapter & Templates & 5.68 & 7.1 & 19.7 & 41.6 & 399k & 44k \\
template adapter & LLM & 5.68 & 7.2 & 20.5 & 43.5 & 442k & 44k \\
LLM adapter & Templates & 7.88 & 9.2 & 21.6 & 43.0 & 531k & 60k \\
LLM adapter & LLM & 7.88 & 9.2 & 21.5 & 42.9 & 521k & 60k \\
Unguided base & Templates & 0.00 & 7.9 & 79.2 & 202.8 & -- & -- \\
Unguided base & LLM & 0.00 & 7.9 & 79.2 & 202.8 & -- & -- \\
\midrule
\multicolumn{8}{@{}l}{\emph{SD-Turbo}} \\
released & Templates & 0.00 & 0.4 & 4.0 & 10.4 & -- & -- \\
released & LLM & 0.00 & 0.4 & 4.1 & 10.5 & -- & -- \\
base-first & Templates & 0.00 & 0.4 & 4.2 & 10.8 & -- & -- \\
base-first & LLM & 0.00 & 0.4 & 4.2 & 10.6 & -- & -- \\
template adapter & Templates & 0.90 & 1.3 & 4.5 & 10.3 & 807k & 38k \\
template adapter & LLM & 0.90 & 1.3 & 4.6 & 10.5 & 1076k & 39k \\
LLM adapter & Templates & 1.16 & 1.5 & 4.8 & 10.6 & 1072k & 50k \\
LLM adapter & LLM & 1.16 & 1.5 & 4.9 & 10.8 & 1442k & 50k \\
unguided base & Templates & 0.00 & 1.5 & 15.4 & 39.4 & -- & -- \\
unguided base & LLM & 0.00 & 1.5 & 15.4 & 39.4 & -- & -- \\
\bottomrule
\end{tabular}

  \caption{\textbf{Cost estimates across dataset sizes from ImageNet-100 throughputs.}
  H100-hours for preparation plus generation, using ImageNet-100 timings with the indicated prompts,
  constant throughput, and one preparation; 1.28M denotes 1,281,167 images. The ImageNet-1k panels in
  \cref{tab:supp-deployment} instead use throughputs measured on the full-scale sets and 1,281,000 images.
  $N^*$ is the approximate image count at which an adapter becomes cheaper than base-first (BF) or the unguided
  base. Classifier evaluation and sampler development are excluded. Unguided-base LLM rows reuse template
  throughput. --: crossover not applicable.}
  \label{tab:supp-e50-costs}
\end{minipage}\par\medskip

\subsection{Generation and adapter implementation}
\paragraph{Set manifest.} The accompanying machine-readable \texttt{set-manifest.json} fixes the 186 template-prompt
sets in the metric study: model and partner, guidance/stage plans, noise schedules, resolution, batch where recorded,
prompt/seed rules, classifier seeds, and accuracy/cost observations. The fixed metric-study list is not expanded
with later experiments. Accompanying source files provide CSV/JSON result snapshots, per-class accuracies,
and scripts to regenerate the rich-prompt, cost, and supporting-contrast tables. The code records the
underlying generation configurations and full run results.

\paragraph{Generator sources.} SD~1.5 and SD~2.1 use latent diffusion~\cite{rombach2022high}; SD-Turbo uses adversarial distillation~\cite{sauer2024adversarial}.
We evaluate SDXL~\cite{podell2024sdxl} with Turbo, DMD2, Lightning, and LCM students~\cite{sauer2024adversarial,yin2024improved,lin2024sdxl,luo2023latent}; FLUX.1-dev and SRPO~\cite{bfl2024flux,shen2025directly}; FLUX.2-klein 4B~\cite{bfl2026klein}; and Z-Image/Z-Image-Turbo~\cite{zimageteam2025zimage}.
The DP-DMD comparison uses SD3.5-Medium~\cite{esser2024scaling,wu2026diversity}.

\paragraph{Prompts and seeds.} Each image of class $k$ uses template $j \bmod 8$ of ``a photo of a \{\}.'', ``a
close-up photo of a \{\}.'', ``a photo of the \{\} in its natural setting.'', ``a bright photo of a \{\}.'', ``a photo of
a small \{\}.'', ``a photo of a large \{\}.'', ``a cropped photo of a \{\}.'' and ``a good photo of a \{\}.'', filled with
the class's first WordNet synonym (a short list of ambiguous names is replaced, \eg ``crane bird'', ``movie theater''),
and the same image seed across generators. Unless otherwise stated, images are generated at 512 pixels, except SDXL and its DMD2, LCM and
Lightning students at their native 1024 (SDXL-Turbo at its native 512; the flow models, native at 1024, at 512), and
stored with their short side at 256 pixels. Every controlled pair thus shares its resolution except SDXL against
SDXL-Turbo; resolution itself changes training value~\cite{azizi2023synthetic}, so that one gap is not read as
distillation alone.

\paragraph{Adapter.} Rank-32 low-rank updates ($\alpha=32$) on the linear layers of the distilled model's transformer
blocks (on the SD U-Net, the linear layers of its attention blocks), trained with AdamW at $10^{-4}$, 100 warmup steps and cosine decay,
3,000 steps of 8 pairs (16 for SD-Turbo), the low-rank part kept and applied in float32. The teacher state $x^\star_{\sigma_a}$ is obtained with the
unguided base's own sampler on a fine grid that contains $\sigma_a$; the pairs use 160 seeds per ImageNet-100 class
name, cycling through the eight templates, and none of these seeds is used for an evaluated set.

\paragraph{SD-Turbo parameterization.} Equation~\ref{eq:adapter} uses flow-matching coordinates. For SD-Turbo,
we use its Euler scheduler's noise coordinate $s(t)=\sqrt{(1-\bar\alpha_t)/\bar\alpha_t}$ and latents
$z_s=z_0+s\epsilon$, where $z_0$ is the clean latent and $\bar\alpha_t=\prod_{u\leq t}(1-\beta_u)$. Write
$s_{\max}=s(999)$ and $s_a=s(499)$. Starting from $z_{\max}=s_{\max}\epsilon$ with
$\epsilon\sim\mathcal{N}(0,I)$, the adapted prediction is
\begin{equation}
  \begin{aligned}
  \hat\epsilon &= \epsilon_{\theta+\Delta W}
  \!\left(\frac{z_{\max}}{\sqrt{1+s_{\max}^2}},999,c\right),\\
  \hat z_a &= z_{\max}+(s_a-s_{\max})\,\hat\epsilon,\\
  \mathcal{L}_{\rm SD} &= \mathbb{E}\,\|\hat z_a-z_a^\star\|^2 .
  \end{aligned}
  \label{eq:sd-adapter}
\end{equation}
Here $\epsilon_\theta$ is SD-Turbo's conditional noise-prediction U-Net. To obtain $z_a^\star$, we run the
unguided SD~2.1 U-Net in the same Euler pipeline, sharing the initial noise, text encoder, and VAE. The fine grid
uses 30 equally spaced timestep intervals from 999 to 499 and 15 from 499 to 249, rounding timesteps to integers;
we stop at the selected target. At inference, the adapted call at 999 jumps directly to $s(499)$. We disable the
update and continue the released student at timesteps 499 and 249, ending at zero noise: three U-Net calls total.

\paragraph{SDA comparison.} We evaluate the released \texttt{zit\_sda\_v1.safetensors} adapter~\cite{fok2026sdazimage} on Z-Image-Turbo at its documented eight steps. Applying it throughout reaches \tpsd{zt_sda_all}, versus \tpsd{zt_sda_first2} with activation limited to the first two steps and \tpsd{zimage_turbo_s8} without it (three classifier seeds each). We use the better activation schedule in the main-text comparison. Our implementation applies its full-rank LoKr in float32 without merging, to preserve small updates; its measured runtime is therefore excluded from the main cost plot.

\paragraph{Baselines.} FSG~\cite{bhat2026settle} follows its published FLUX.2-klein hyperparameters; its update as
written does not move klein's block-1 features at its learning rate (with feature norms of about 2,000, the
update falls below float32 resolution), so we take the same steps with Adam. MoNO~\cite{shi2026manifold} rotates the low
frequencies of each initial noise against the DINOv2 features of the earlier images of its class, with four images
per basis as published or 32. Interval guidance~\cite{kynkaanniemi2024applying} uses the interval published for
SD-XL, EDM noise levels $(0.28,5.42]$, at each model's default guidance weight.

\subsection{Data provenance and impact}
 We use existing recognition datasets, released generators, and COCO captions; we do not collect a new human-subject dataset. Generated data can inherit the source models' biases and omit rare visual conditions. Higher aggregate accuracy or recall does not establish demographic fairness or safety in deployment; downstream uses require validation on their intended populations and conditions. The adapter reduces generation cost relative to full-step sampling, but target construction and classifier evaluation add computation. Preparation and per-image generation costs are reported separately.

\FloatBarrier\section{Evidence for the teacher target and update placement}
\subsection{Early guidance and base/student handoffs}
\label{sec:supp-trajectory}
\par\medskip\noindent
\begin{minipage}{\linewidth}
  \captionsetup{type=figure,hypcap=false}
  \centering
  \begin{subfigure}[t]{0.49\linewidth}
    \includegraphics[width=\linewidth]{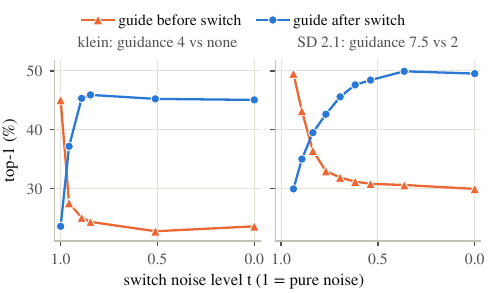}
    \caption{Guidance only above or only below a noise level.}
    \label{fig:window}
  \end{subfigure}\hfill
  \begin{subfigure}[t]{0.49\linewidth}
    \includegraphics[width=\linewidth]{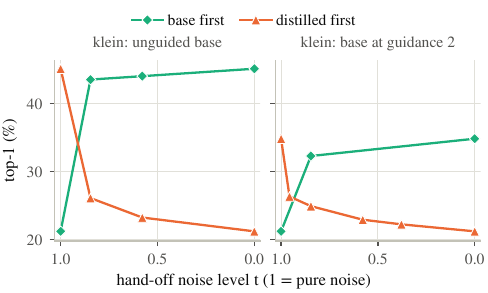}
    \caption{Hand-offs between the distilled model and the base.}
    \label{fig:handoff}
  \end{subfigure}
  \caption{\textbf{Early guidance and base predictions have larger effects on classifier accuracy.}
ImageNet-100 ResNet-50 top-1. (a) Guidance applied only above or below a noise
level: klein at guidance 4 versus none, SD~2.1 at 7.5 versus 2. Early-only guidance incurs most of the loss;
late-only guidance preserves most of the value. (b) Handoffs between the base and student on klein's 50-step base schedule,
with the base unguided (left) or at guidance 2 (right). Base-first uses the base above the switch level and the
student below it; base-last reverses them. Early base steps restore value that late base steps cannot.}
  \label{fig:trajectory}
\end{minipage}\par\medskip
\paragraph{Guidance windows locate the loss.} Guiding only the first 10 of 50 FLUX.2-klein steps costs
\gpa{hiTenKlein} of the \gpa{guideKlein} points lost with full-trajectory guidance. Guiding only after step 20,
below noise 0.89, preserves unguided value (\gp{loTwentyKlein}; \cref{fig:window}). For SD~2.1, guidance 7.5 in the
first 9 of 30 steps, followed by guidance 2, costs \gpa{hiNineSD} of the \gpa{guideSDTwo} points lost with guidance
7.5 throughout, relative to guidance 2. These comparisons locate much of the measured loss in the early sampling steps.

\paragraph{Correcting the beginning restores value; correcting the end does little.} On the base's 50-step
schedule, unguided klein followed by its student at noise 0.85 reaches \tp{kb_b2d25_g1}; reversing their order
reaches \tp{kb_d2b25_g1} (\cref{fig:handoff}). On the distilled four-step schedule, replacing the first two
predictions with the unguided base raises klein from \tp{klein} to \tp{kb_dd2_g1} and SD-Turbo from \tp{sdt4} to
\tp{sdt4_dd2_g1}. This extends the diversity asymmetry of~\cite{gandikota2026distilling} to training value.

Completing a fine-grid base state at the student's first level reaches \tp{kb_or10_d3} on klein and
\tp{sdt_or15_d3} on SD-Turbo; stopping at its second level reaches \tp{kb_or21_d2} and \tp{sdt_or30_d2}.
These comparisons identify an early interval where replacing student predictions changes classifier accuracy for
the specified generator, prompts, and student completion. Successful recovery requires compatibility with the
remaining student trajectory (\cref{sec:adapter}).

\paragraph{Diversity-oriented settings do not guarantee recovery.} Published interval
guidance~\cite{kynkaanniemi2024applying}, transferred without retuning, reaches \tp{kb_interval_g4} on klein but
only \tp{sd21_interval75} on SD~2.1, versus \tp{sd21_g1} unguided. The transferred interval still applies guidance within
SD~2.1's sensitive early range, illustrating why a guidance schedule's effect on training value depends on
which noise levels it covers. We report the transferred interval settings in \cref{sec:supp-protocol}.

\subsection{Target design and stability}

\label{sec:supp-additional}

\paragraph{Stability across adapter seeds.} Each primary adapter is trained three times. ImageNet-100 accuracy
ranges are \nm{adKleinRange}, \nm{adSDRange}, and \nm{adZRange} for klein, SD-Turbo, and Z-Image, respectively.

\paragraph{Independent dataset regeneration.}
\Cref{tab:supp-regeneration} compares the four headline sets with one fresh generation of each, using new
initial noise and unchanged sampler settings, prompts, and adapter weights. The largest change in mean
accuracy is \nm{regenMax} points, for the released SD-Turbo; the adapter's advantage remains large in both
families. Each row contains two generated datasets evaluated with three classifier seeds apiece. This check
separates dataset regeneration from classifier initialization, but two realizations do not estimate the full
variance over generated datasets.

\par\medskip\noindent
\begin{minipage}{\linewidth}
  \captionsetup{type=table,hypcap=false}
  \centering
  \tablebody
\begin{tabular}{@{}lrr@{}}
\toprule
Sampler & Original set & Regenerated set \\
\midrule
FLUX.2-klein, released & 17.1\,{\tableuncert$\pm$0.4} & 17.2\,{\tableuncert$\pm$0.4} \\
FLUX.2-klein + adapter & 44.1\,{\tableuncert$\pm$0.5} & 44.4\,{\tableuncert$\pm$0.5} \\
SD-Turbo, released & 18.3\,{\tableuncert$\pm$0.3} & 17.1\,{\tableuncert$\pm$0.7} \\
SD-Turbo + adapter & 43.3\,{\tableuncert$\pm$0.2} & 43.9\,{\tableuncert$\pm$0.1} \\
\bottomrule
\end{tabular}

  \caption{\textbf{The adapter gains persist after regenerating the training images.}
  ImageNet-100 ResNet-50 top-1 (mean $\pm$ sd over three classifier seeds for each generated set), using
  500 images per class. Original and regenerated sets use different initial noise with the same prompt
  construction. Adapters use fixed weights from their first training run; variation across these two datasets
  is distinct from the classifier-seed standard deviations shown.}
  \label{tab:supp-regeneration}
\end{minipage}\par\medskip

\paragraph{Guided targets forfeit much of the gain.} With the base's default guidance applied at every target
integration step, as in DP-DMD's released teacher implementation, the adapter loses \gpa{tgFourKlein} points
(\gpcia{tgFourKlein}) on klein and \gpa{tgSevenSD} (\gpcia{tgSevenSD}) on SD-Turbo. It retains only
\nm{tgShareKlein}\% and \nm{tgShareSD}\% of the unguided-target gain. Completing the guided teacher states with
the distilled model gives \tp{kb_or35_d1_g4} and \tp{sdt_or30_d2_g75}; the guided-target adapters perform at or
below those references. On 200 classes, their losses relative to the standard adapters are
\gpa{tgFourKleinTin} and \gpa{tgSevenSDTin} points.

The same direction holds on Z-Image: the default-guidance target costs \gpa{tgZ} (\gpcia{tgZ}) points,
\gpa{tgZTin} on 200 classes, and retains \nm{tgShareZ}\% of the gain. This is consistent with 1.x-Distill's motivation for removing
high-noise teacher guidance to preserve layouts~\cite{li2026distill}, but evaluates a different outcome.
In all three families, the default-guidance targets also retain at most \nm{tgVendiMax}\% of the Vendi gain and
\nm{tgRecallMax}\% of the recall gain (first training runs). Thus the guided targets lose diversity and training value together.

\paragraph{The target must span the consequential early interval.} Ending at the student's original first level
reaches only \tp{kd_ad0} on klein and \tp{sdt_ad0} on SD-Turbo, retaining \nm{anchorShareKlein}\% and
\nm{anchorShareSD}\% of the full gain. Those results resemble completion from the base's own states at the same
levels (\tp{kb_or10_d3} and \tp{sdt_or15_d3}); a later distilled step still traverses the sensitive interval.
Vendi gains are similarly limited (\nm{anchorVendiKlein}\% and \nm{anchorVendiSD}\%). Deeper is not always better:
\tp{kd_ad3} and \tp{sdt_ad2} for deeper klein and SD-Turbo jumps fall below the selected targets. These comparisons support the selected intermediate target over either the original first endpoint or the
deepest tested endpoint; they do not define a universal structural threshold.

\paragraph{Seedwise target correspondence.}
The adapter preserves target content for individual seeds (\cref{fig:method}). In DINOv3 space, an adapted
image is more similar to the completed teacher state from its own initial noise than to states from other noise.
The cosine-similarity advantage is \nm{relayAdKlein} for klein, compared with \nm{relayBfKlein} for base-first
and \nm{relayDistKlein} for the distilled model. Corresponding SD-Turbo values are \nm{relayAdSD},
\nm{relayBfSD}, and \nm{relayDistSD}. This correspondence supports the interpretation that the adapter learns
the target trajectory while the aggregate diversity comparisons alone do not identify how that correspondence arises.

\FloatBarrier\section{Additional baseline and operating-point comparisons}
\subsection{Guidance and distillation across generator families}
\label{sec:attribution}
The primary families' guided, unguided, and distilled references are reported numerically in
\cref{tab:supp-main-complete}. \Cref{tab:supp-secondary-generators} adds SD~1.5, SDXL, and FLUX.1-dev without
redrawing the main comparison.

\par\medskip\noindent
\begin{minipage}{\linewidth}
  \captionsetup{type=table,hypcap=false}
  \centering
  \tablebody
  \setlength{\tabcolsep}{3pt}
  \renewcommand{\arraystretch}{1.06}
\begin{tabular}{@{}llrrr@{}}
\toprule
Family & Setting & Steps & Pixels & Top-1 (\%) \\
\midrule
SD 1.5 & Base, guidance 7.5 & 30 & 512 & 32.9\,{\tableuncert$\pm$0.5} \\
 & Base, guidance 2 & 30 & 512 & \textbf{52.1}\,{\tableuncert$\pm$0.4} \\
\midrule
SDXL & Base, guidance 5 & 30 & 1024 & 28.6\,{\tableuncert$\pm$0.3} \\
 & Base, guidance 2 & 30 & 1024 & \textbf{32.0}\,{\tableuncert$\pm$1.3} \\
 & Turbo & 1 & 512 & 12.9\,{\tableuncert$\pm$0.5} \\
 & DMD2 & 4 & 1024 & 18.1\,{\tableuncert$\pm$0.4} \\
 & Lightning & 4 & 1024 & 20.2\,{\tableuncert$\pm$0.2} \\
 & LCM, guidance 8 & 4 & 1024 & 18.1\,{\tableuncert$\pm$0.2} \\
\midrule
FLUX.1-dev & Guidance 3.5 & 28 & 512 & 12.6\,{\tableuncert$\pm$0.3} \\
 & Guidance 1 & 28 & 512 & \textbf{30.6}\,{\tableuncert$\pm$0.5} \\
 & SRPO, guidance 3.5 & 28 & 512 & 15.7\,{\tableuncert$\pm$0.3} \\
\bottomrule
\end{tabular}

  \caption{\textbf{Guidance and post-training effects in additional generator families.}
  ImageNet-100 ResNet-50 top-1 (mean $\pm$ sd over three classifier seeds), using class-name templates.
  Steps and generation resolution are shown for every setting; stored images are resized as in
  \cref{sec:supp-protocol}. Bold: best tested setting per family. SDXL-Turbo's different resolution prevents
  attributing its entire gap to distillation.}
  \label{tab:supp-secondary-generators}
\end{minipage}\par\medskip

\paragraph{Guidance settings are model-specific.} Removing guidance improves the tested klein and Z-Image
bases, while reducing guidance to 2 improves SD~1.5, SD~2.1, and SDXL. SD~2.1 reaches \tp{sd21_g2}, compared
with \tp{sd21} at guidance 7.5. For FLUX.1-dev, the conditioning input encodes distilled guidance rather than
an additional unconditional model evaluation. Its change from 3.5 to 1 also improves classifier accuracy.
These comparisons support changing guidance in the studied configurations; they do not identify one
universally optimal weight.

\paragraph{Post-training need not reduce training value.} The evaluated distilled releases perform at or
below their tested guided bases. The four SDXL students in \cref{tab:supp-secondary-generators} cover
adversarial, distribution-matching, progressive, and consistency objectives~\cite{sauer2024adversarial,yin2024improved,lin2024sdxl,luo2023latent}.
Three use the base's resolution, so resolution alone cannot explain that pattern. By contrast, SRPO improves
on FLUX.1-dev by \gp{srpo} (\cref{tab:supp-contrasts}). The observed deficit is therefore not an inevitable
consequence of generator post-training.

\subsection{Additional generation-cost comparisons}
\par\medskip\noindent
\begin{minipage}{\linewidth}
  \captionsetup{type=figure,hypcap=false}
  \centering
  \includegraphics[width=\linewidth]{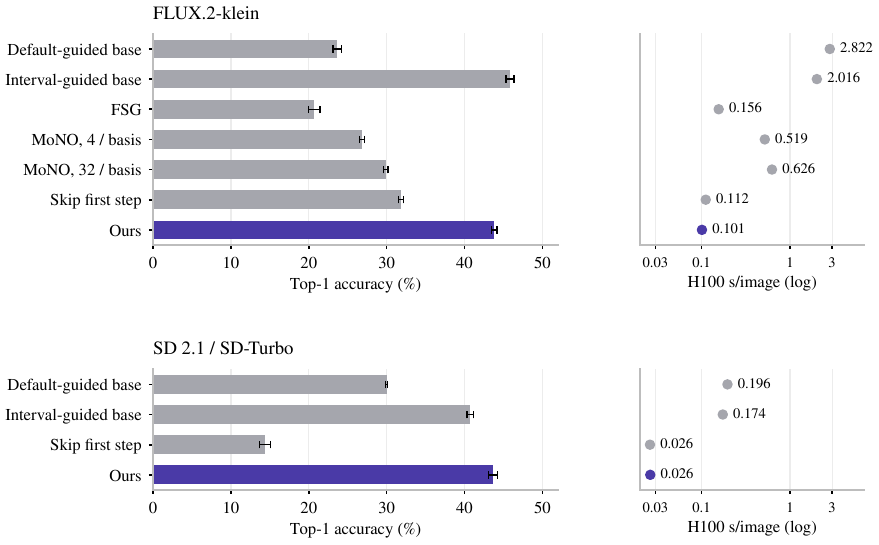}
  \caption{\textbf{Additional baselines offer different accuracy and cost tradeoffs.}
  ImageNet-100 ResNet-50 top-1 (left, mean $\pm$ sd) and H100 seconds/image (right, logarithmic scale).
  Adapters pool three training runs with three classifier seeds each; other settings use three classifier seeds.
  Default and interval guidance apply to the base; FSG, MoNO, and skipping the first step apply to the released
  student. MoNO uses four or 32 earlier images per basis. Ours is repeated as a reference; released-student,
  base-first, and unguided-base comparisons are in \cref{fig:cost}. Times include the training-free methods'
  optimization but exclude adapter preparation (\cref{tab:supp-deployment}); implementation details are in
  \cref{sec:supp-protocol}.}
  \label{fig:supp-cost-full}
\end{minipage}\par\medskip
\Cref{fig:supp-cost-full} isolates the operating points omitted from the main cost figure. Transferred interval
guidance performs well on klein but remains below unguided sampling on SD~2.1. FSG, both MoNO basis sizes, and
skipping the first step remain below the klein adapter. Skipping the first SD-Turbo step does not improve on
the released student. These results concern the transferred implementations and settings described in
\cref{sec:supp-protocol}, not optimized limits for each method.

\subsection{Additional primary results and alternative adaptation prompts}
\Cref{tab:supp-main-complete} reports additional primary comparisons, including inference costs and adapters
fitted using COCO captions instead of class-name templates. The additional unguided-base references for
the 200-class and full ImageNet-1k protocols are reported in \cref{tab:main}.
\par\medskip\noindent
\begin{minipage}{\linewidth}
  \captionsetup{type=table,hypcap=false}
  \centering
  \small
  \renewcommand{\arraystretch}{1.02}
\begin{tabular}{@{}llrrrrr@{}}
\toprule
family & sampler & NFE & s/img & ImageNet-100 & 200 classes & ImageNet-1k \\
\midrule
\multirow{7}{*}{FLUX.2-klein} & distilled, 4 steps & 4 & 0.14 & 17.1\,{\tableuncert$\pm$0.4} & 19.1\,{\tableuncert$\pm$0.0} & 8.4\,{\tableuncert$\pm$0.1} \\
 & base-first & 4 & 0.15 & 39.9\,{\tableuncert$\pm$0.3} & 43.7\,{\tableuncert$\pm$0.1} & 30.8\,{\tableuncert$\pm$0.1} \\
 & + adapter (ours) & 2 & 0.10 & 43.8\,{\tableuncert$\pm$0.4} & \textbf{48.6}\,{\tableuncert$\pm$0.3} & \textbf{34.2}\,{\tableuncert$\pm$0.1} \\
 & \quad trained on COCO captions & 2 & 0.10 & \textbf{45.0}\,{\tableuncert$\pm$0.5} & 48.5\,{\tableuncert$\pm$0.2} & -- \\
 & base, no guidance & 50 & 1.38 & 45.1\,{\tableuncert$\pm$0.0} & 48.2\,{\tableuncert$\pm$0.1} & -- \\
 & base, no guidance, 20 steps & 20 & 0.57 & 44.7\,{\tableuncert$\pm$0.9} & 48.1\,{\tableuncert$\pm$0.2} & 37.2\,{\tableuncert$\pm$0.1} \\
 & base, default guidance & 100 & 2.82 & 23.6\,{\tableuncert$\pm$0.5} & 27.0\,{\tableuncert$\pm$0.2} & -- \\
\midrule
\multirow{6}{*}{SD-Turbo} & distilled, 4 steps & 4 & 0.03 & 18.3\,{\tableuncert$\pm$0.3} & 19.8\,{\tableuncert$\pm$0.3} & 12.7\,{\tableuncert$\pm$0.2} \\
 & base-first & 4 & 0.03 & 36.4\,{\tableuncert$\pm$0.9} & 36.0\,{\tableuncert$\pm$0.2} & 26.7\,{\tableuncert$\pm$0.3} \\
 & + adapter (ours) & 3 & 0.03 & \textbf{43.7}\,{\tableuncert$\pm$0.6} & \textbf{41.3}\,{\tableuncert$\pm$0.8} & \textbf{32.1}\,{\tableuncert$\pm$0.3} \\
 & \quad trained on COCO captions & 3 & 0.03 & 42.4\,{\tableuncert$\pm$1.3} & 41.0\,{\tableuncert$\pm$1.0} & -- \\
 & base, no guidance & 30 & 0.11 & 49.8\,{\tableuncert$\pm$0.7} & 45.7\,{\tableuncert$\pm$0.9} & 44.1\,{\tableuncert$\pm$0.1} \\
 & base, default guidance & 60 & 0.20 & 30.0\,{\tableuncert$\pm$0.1} & 31.5\,{\tableuncert$\pm$0.3} & -- \\
\midrule
\multirow{5}{*}{Z-Image-Turbo} & distilled, 9 steps & 9 & 0.37 & 14.0\,{\tableuncert$\pm$0.0} & 13.9\,{\tableuncert$\pm$0.4} & -- \\
 & base-first & 9 & 0.37 & 32.8\,{\tableuncert$\pm$0.9} & 35.3\,{\tableuncert$\pm$0.3} & -- \\
 & + adapter (ours) & 7 & 0.30 & \textbf{35.1}\,{\tableuncert$\pm$0.5} & \textbf{38.2}\,{\tableuncert$\pm$0.3} & -- \\
 & base, no guidance & 50 & 1.92 & 45.6\,{\tableuncert$\pm$0.7} & -- & -- \\
 & base, default guidance & 100 & 3.78 & 17.6\,{\tableuncert$\pm$0.3} & -- & -- \\
\bottomrule
\end{tabular}

  \caption{\textbf{Adapting the first prediction improves classifier accuracy at student sampling cost.}
Synthetic-only ResNet-50 top-1 (\%). Primary ImageNet-100 adapters pool three training runs; other adapter results use one. Columns: ImageNet-100 and 200-class development (186 classes unseen during adaptation),
and full ImageNet-1k with fixed samplers; images/class:
500, 500, and 1,281. Indented: adapters trained on 16,000 COCO
captions instead of class-name templates. NFE: denoiser evaluations; s/img: H100 seconds/image.
Bold: best fast sampler per family and protocol.}
  \label{tab:supp-main-complete}
\end{minipage}\par\medskip

\subsection{DP-DMD's released student}

\label{sec:supp-dpdmd}

\par\medskip\noindent
\begin{minipage}{\linewidth}
  \captionsetup{type=table,hypcap=false}
  \centering
  \small
  \renewcommand{\arraystretch}{1.06}
\begin{tabular}{@{}lrrrrr@{}}
\toprule
Sampler & Pixels & NFE & s/img & ImageNet-100 & 200 classes \\
\midrule
SD3.5-Medium, guidance 3.5 & 512 & 56 & 0.613 & 21.1\,{\tableuncert$\pm$0.5} & 25.1\,{\tableuncert$\pm$0.4} \\
SD3.5-Medium, no guidance & 512 & 28 & 0.326 & 44.9\,{\tableuncert$\pm$0.3} & 44.7\,{\tableuncert$\pm$0.4} \\
DP-DMD's student & 512 & 4 & 0.071 & 13.9\,{\tableuncert$\pm$0.2} & 15.3\,{\tableuncert$\pm$0.3} \\
\quad base-first, first level & 512 & 4 & 0.091 & 15.8\,{\tableuncert$\pm$0.1} & 17.5\,{\tableuncert$\pm$0.4} \\
\quad + adapter, first level & 512 & 4 & 0.083 & 21.6\,{\tableuncert$\pm$0.3} & \textbf{24.2}\,{\tableuncert$\pm$0.1} \\
\quad + adapter, second level & 512 & 3 & 0.072 & \textbf{27.3}\,{\tableuncert$\pm$0.2} & -- \\
\midrule
DP-DMD's student & 1024 & 4 & 0.266 & 15.8\,{\tableuncert$\pm$0.6} & -- \\
\quad + adapter, second level & 1024 & 3 & 0.238 & \textbf{27.2}\,{\tableuncert$\pm$0.6} & -- \\
\bottomrule
\end{tabular}

  \caption{\textbf{Our adapter on DP-DMD's released SD3.5-Medium student.} Top-1 (\%, mean $\pm$ sd over three
  classifier seeds; the first-level adapter on ImageNet-100 pools three training runs) of a ResNet-50 trained from
  scratch on 500 generated images per class. The student runs its own four-step schedule at guidance 1; its first
  level is 0.858 and its second 0.602. Base-first uses one unguided teacher prediction to the first student
  level, followed by three student predictions; adapter targets use fine-grid teacher integration. All sets at
  512 pixels in the first block, 1024 in the last two rows; the same adapter weights are used at both resolutions. NFE and s/img as in \cref{tab:supp-main-complete}. Bold: best fast sampler per protocol.}
  \label{tab:supp-dpdmd}
\end{minipage}\par\medskip

DP-DMD~\cite{wu2026diversity} regresses its student's first step onto the teacher's state at guidance 3.5 and trains
the later steps by distribution matching. Its released SD3.5-Medium student falls \gpa{dpdmdVsTeacher} points below
the guided teacher (\gpcia{dpdmdVsTeacher}), while removing guidance raises the teacher by \gpa{dpdmdGuide} points
(\cref{tab:supp-dpdmd}). Our adapter, fitted on top of the student toward the unguided teacher, adds \gp{dpdmdAd}
(\gpci{dpdmdAd}) at the student's first level and \gp{dpdmdAdDeep} (\gpci{dpdmdAdDeep}) at its second, and exceeds
base-first at the first level by \gp{dpdmdAdVsBf} (\gpci{dpdmdAdVsBf}; \gp{dpdmdAdVsBfTin} on 200 classes). It
recovers \nm{dpdmdAdShare}\% and \nm{dpdmdAdDeepShare}\% of what the unguided teacher adds over the student.
Base-first reaches \tp{dpdmd_dd1_g1} after one unguided teacher prediction to the first student level and three
student predictions. Our adapter instead learns from a teacher state obtained by fine-grid integration.
The comparison therefore establishes an advantage over this coarse base-first sampler, but does not isolate
the effects of teacher integration, hand-off mismatch, and student continuation.

\paragraph{Documented resolution.} At DP-DMD's documented 1024-pixel resolution, the unchanged adapter fitted
at 512 pixels adds \gp{e49native} ImageNet-100 points (\gpci{e49native}; \cref{tab:supp-dpdmd}).
The student's noise schedule is unchanged because its scheduler uses a static shift. Increasing resolution
changes the released student's accuracy by \gp{e49resolutionReleased} points (\gpci{e49resolutionReleased}) and
the adapted student's by \gp{e49resolutionAdapted} (\gpci{e49resolutionAdapted}). No adapter weights or sampling
endpoints are adjusted for this comparison.

\subsection{Relation to earlier distillation and adaptation methods}
\par\medskip\noindent
\begin{minipage}{\linewidth}
  \captionsetup{type=table,hypcap=false}
  \centering
  \footnotesize
  \setlength{\tabcolsep}{3pt}
  \renewcommand{\arraystretch}{1.15}
  \begin{tabular}{@{}p{.11\linewidth}p{.12\linewidth}p{.235\linewidth}p{.185\linewidth}p{.085\linewidth}p{.17\linewidth}@{}}
\toprule
\raggedright Method & \raggedright Starting point & \raggedright Supervision & \raggedright Trained sampling steps & \raggedright Base at inference & \raggedright Reported evaluation \tabularnewline
\midrule
\raggedright DP-DMD~\cite{wu2026diversity} & \raggedright Teacher-to-student distillation & \raggedright Teacher-derived first-step target; distribution matching afterward & \raggedright First and remaining steps & \raggedright No & \raggedright Image quality and diversity \tabularnewline
\addlinespace
\raggedright CrossDistill~\cite{liu2026crossdistill} & \raggedright Teacher-to-student distillation & \raggedright High-noise trajectory preservation; low-noise distribution matching & \raggedright High- and low-noise stages & \raggedright No & \raggedright Video quality and diversity \tabularnewline
\addlinespace
\raggedright SDA~\cite{fok2026sdazimage} & \raggedright Released image student & \raggedright Teacher-guided directional alignment with self-reference & \raggedright Adapter trained across sampling noise levels & \raggedright No & \raggedright Image diversity \tabularnewline
\addlinespace
\raggedright Ours & \raggedright Released image student & \raggedright Regression to an unguided intermediate teacher state & \raggedright First prediction; remaining steps unchanged & \raggedright No & \raggedright Synthetic-to-real classifier accuracy \tabularnewline
\bottomrule
\end{tabular}

  \caption{\textbf{First-step supervision and post-hoc adaptation are precedents for our procedure.}
  CrossDistill's direct-flow variant regresses the first prediction to an intermediate teacher state;
  its default uses consistency supervision. The SDA row describes the Z-Image-Turbo release evaluated here;
  adapter activation schedules differ across SDA releases. All four methods generate without the base at
  inference. Our comparisons test downstream classifier accuracy when adapting a released student while
  retaining its unchanged completion steps. Evaluation columns summarize the cited releases' reported tests,
  not the possible uses of each method.}
  \label{tab:supp-method-positioning}
\end{minipage}\par\medskip

\FloatBarrier\section{Dataset scale and learner generalization}
\subsection{Dataset scale and additional classes}

\label{sec:supp-scaling}

\paragraph{Full-scale transfer beyond adapter-training classes.} Restricting the saved full-scale ImageNet-1k
evaluation to the 900 classes outside ImageNet-100 retains the large adapter gains and the adapter's advantage
over base-first (\cref{tab:supp-unseen-fullscale}).
We keep the original 1,000-way predictions and average per-class accuracies over those 900 classes, rather than
restrict the classifier's output space. Each classifier was trained on generated images of all 1,000 classes;
the adapter was trained on ImageNet-100 prompts only. All 1,000 classes appeared in the earlier scale pilot;
this subgroup evaluates fixed-setting scale transfer beyond adapter fitting. The accuracy-gap fractions reported in \cref{sec:experiments} use
$(A_{\rm adapter}-A_{\rm student})/(A_{\rm base}-A_{\rm student})$, with $A$ the accuracy on the matched protocol
and the base sampled without guidance.

\par\medskip\noindent
\begin{minipage}{\linewidth}
  \captionsetup{type=table,hypcap=false}
  \centering
  \tablebody
  \setlength{\tabcolsep}{3pt}
\begin{tabular}{@{}lrrrrr@{}}
\toprule
Generator & Released & Base-first & + adapter & Over released [95\% CI] & Over base-first [95\% CI] \\
\midrule
FLUX.2-klein & 8.6\,{\tableuncert$\pm$0.1} & 31.0\,{\tableuncert$\pm$0.1} & \textbf{34.6}\,{\tableuncert$\pm$0.1} & \shortstack{+26.0\\{\tableuncert[+24.8, +27.3]}} & \shortstack{+3.6\\{\tableuncert[+3.0, +4.3]}} \\
SD-Turbo & 12.9\,{\tableuncert$\pm$0.2} & 26.8\,{\tableuncert$\pm$0.3} & \textbf{32.2}\,{\tableuncert$\pm$0.3} & \shortstack{+19.3\\{\tableuncert[+18.3, +20.3]}} & \shortstack{+5.4\\{\tableuncert[+4.9, +5.8]}} \\
\bottomrule
\end{tabular}

  \caption{\textbf{Full-scale gains persist outside the adapter's training classes.}
  ImageNet-1k top-1 (\%) on the 900 classes outside ImageNet-100, keeping all 1,000 classifier outputs. Mean $\pm$
  SD over three classifier seeds, one adapter run. Gaps compare the adapter with each reference using paired
  95\% class-bootstrap intervals after averaging classifier runs; intervals condition on the observed datasets
  and fitted checkpoints. Bold: best shown per family. Training uses 1,281 generated images per class for all 1,000 classes.}
  \label{tab:supp-unseen-fullscale}
\end{minipage}\par\medskip

\par\medskip\noindent
\begin{minipage}{\linewidth}
  \captionsetup{type=figure,hypcap=false}
  \centering
  \includegraphics[width=0.55\linewidth]{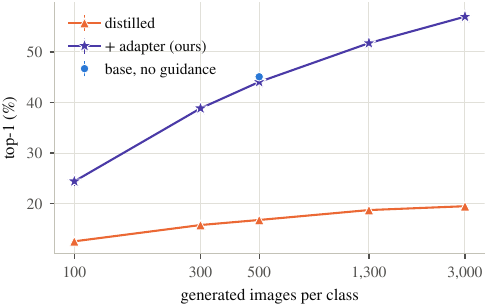}
  \caption{\textbf{Adaptation improves accuracy more than increasing the released model's dataset size.} Top-1 (\%, ImageNet-100)
  of a ResNet-50 trained from scratch on 100 to 3,000 images per class from distilled FLUX.2-klein and from the same
  model with the adapter, 100 epochs at every size, mean $\pm$ sd over two classifier seeds (three for the adapter at
  500). The sizes are nested subsets of one 3,000-per-class set per sampler, with every prompt template equally
  represented. Blue: the unguided base at 500 per class.}
  \label{fig:supp-scaling}
\end{minipage}\par\medskip

Increasing the ImageNet-100 set from 100 to 3,000 images per class raises distilled FLUX.2-klein from \tp{scale_klein_100} to \tp{scale_klein_3000}
(\cref{fig:supp-scaling}); with the adapter, classifier accuracy reaches \tp{scale_kd_ad2_100} with 100 images
per class and \tp{scale_kd_ad2_3000} with 3,000. All sizes use 100 epochs. Larger datasets therefore receive more optimization steps; this comparison
measures data scaling under a fixed-epoch protocol rather than fixed training compute. At ImageNet-1k scale the distilled model's curve is flat: classifier accuracy is
\tp{in1k_klein_full} at 1,281 images per class and \tp{in1k_klein} at 100, while the adapted model's accuracy
rises from \tp{in1k_kd_ad2} to \tp{in1k_kd_ad2_full} (\cref{tab:main}).

At 100 images per class on all 1,000 ImageNet classes, klein's adapter reaches \tp{in1k_kd_ad2}, compared with
\tp{in1k_kb_g1_s20} for the unguided base (\gp{inkAdVsBaseHundred}, \gpci{inkAdVsBaseHundred}). SD-Turbo's adapter
reaches \tp{in1k_sdt_ad1}, compared with \tp{in1k_sd21_g1} for unguided SD~2.1 (\gp{inkSdtAdVsBase},
\gpci{inkSdtAdVsBase}).

\subsection{Additional generator comparisons on fine-grained datasets}
\label{sec:supp-crossdata}
The SDA and DP-DMD comparisons use the same ImageNet-100 adapters and sampling settings on Food-101,
Cars, and Pets. Prompts and image seeds are matched within each dataset. The released Z-Image-Turbo uses
nine steps, SDA eight, and our adapter seven denoiser evaluations; these schedules are identified in
\cref{tab:finegrained}. Adding our adapter to DP-DMD measures an improvement to its released student,
rather than a comparison between independently trained distillation procedures.

\par\medskip\noindent
\begin{minipage}{\linewidth}
  \captionsetup{type=table,hypcap=false}
  \centering
  \tablebody
  \setlength{\tabcolsep}{4pt}
\begin{tabular}{@{}lrrr@{}}
\toprule
Mean-class accuracy gain [95\% CI] & Food-101 & Cars & Pets \\
\midrule
SDA $-$ released Z-Image-Turbo & \textbf{+7.7}\,{\tableuncert[+5.7, +9.6]} & \textbf{+27.1}\,{\tableuncert[+24.9, +29.4]} & \textbf{+22.3}\,{\tableuncert[+18.6, +26.3]} \\
Ours $-$ released Z-Image-Turbo & \textbf{+14.2}\,{\tableuncert[+11.3, +17.0]} & \textbf{+32.4}\,{\tableuncert[+29.8, +34.9]} & \textbf{+33.9}\,{\tableuncert[+27.9, +40.0]} \\
Ours $-$ SDA & \textbf{+6.5}\,{\tableuncert[+4.3, +8.8]} & \textbf{+5.3}\,{\tableuncert[+3.9, +6.7]} & \textbf{+11.6}\,{\tableuncert[+7.6, +15.3]} \\
DP-DMD + ours $-$ DP-DMD & \textbf{+15.0}\,{\tableuncert[+12.6, +17.6]} & \textbf{+10.8}\,{\tableuncert[+9.0, +12.7]} & \textbf{+34.1}\,{\tableuncert[+28.9, +39.3]} \\
\bottomrule
\end{tabular}
\par\medskip
\begin{tabular}{@{}lrrrr@{}}
\toprule
H100 seconds/image & NFE & Food-101 & Cars & Pets \\
\midrule
Z-Image-Turbo, released & 9 & 0.369 & 0.370 & 0.369 \\
+ SDA & 8 & 0.706 & 0.707 & 0.705 \\
+ ours & 7 & 0.295 & 0.294 & 0.293 \\
DP-DMD's student & 4 & 0.071 & 0.072 & 0.071 \\
+ ours & 3 & 0.073 & 0.072 & 0.073 \\
\bottomrule
\end{tabular}

  \caption{\textbf{Our adapter exceeds SDA and improves DP-DMD's student across three datasets.}
  Upper block: mean-class accuracy gains in points [paired 95\% class-bootstrap CI], after averaging three
  classifier seeds; bold gains have intervals above zero. Gaps weight classes equally, while top-1 in
  \cref{tab:finegrained} weights test images. Each arm uses one generated set; adapted arms use one fixed checkpoint.
  Lower block: measured H100 seconds/image at the recorded batch sizes.
  SDA uses an unmerged float32 LoKr, so its runtime is not an optimized deployment comparison.}
  \label{tab:supp-crossdata}
\end{minipage}\par\medskip

All twelve class-bootstrap intervals in \cref{tab:supp-crossdata} lie above zero. Both SDA and our adapter
exceed the released Z-Image-Turbo sampler, our adapter exceeds SDA, and the fixed DP-DMD adapter improves
its released student on each dataset. These intervals condition on the generated sets and fitted adapters;
they do not estimate variation across independent regenerations or adapter-training runs.

\par\medskip\noindent
\begin{minipage}{\linewidth}
  \captionsetup{type=table,hypcap=false}
  \centering
  \tablebody
\begin{tabular}{@{}lrrr@{}}
\toprule
Training images & Food-101 & Cars & Pets \\
\midrule
FLUX.2-klein, COCO targets & 31.8\,{\tableuncert$\pm$0.6} & 33.0\,{\tableuncert$\pm$0.5} & -- \\
FLUX.2-klein, unguided base & 29.8\,{\tableuncert$\pm$0.2} & 33.7\,{\tableuncert$\pm$0.9} & 47.4\,{\tableuncert$\pm$0.3} \\
SD-Turbo, COCO targets & 31.3\,{\tableuncert$\pm$0.7} & 67.8\,{\tableuncert$\pm$0.6} & -- \\
SD 2.1, unguided base & 37.3\,{\tableuncert$\pm$0.2} & 69.6\,{\tableuncert$\pm$0.0} & 65.5\,{\tableuncert$\pm$0.6} \\
Real training images & 79.3\,{\tableuncert$\pm$0.3} & 87.0\,{\tableuncert$\pm$0.2} & 64.1\,{\tableuncert$\pm$0.7} \\
\bottomrule
\end{tabular}

  \caption{\textbf{Additional references and COCO-target adapters for fine-grained recognition.}
  ResNet-50 top-1 (\%, mean $\pm$ SD over three classifier seeds). COCO-target adapters use their first training
  run; unguided bases use 20 steps for klein and 30 for SD~2.1. Generated sets contain 500 images/class.
  Real references use 500 images/class for Food-101, all 8,144 training images for Cars, and all 3,680 for
  Pets, with matched classifier-training steps. --: not evaluated. Primary sampler results are in
  \cref{tab:finegrained}.}
  \label{tab:supp-finegrained-references}
\end{minipage}\par\medskip

\subsection{Transfer of learned representations}
\par\medskip\noindent
\begin{minipage}{\linewidth}
  \captionsetup{type=table,hypcap=false}
  \centering
  \tablebody
  \setlength{\tabcolsep}{3pt}
  \renewcommand{\arraystretch}{1.02}
\begin{tabular}{@{}lrrrr@{}}
\toprule
training source & Food-101 & Cars & Pets & IN-900 \\
\midrule
\multicolumn{5}{@{}l}{\emph{FLUX.2-klein}} \\
distilled & 31.9\,{\tableuncert$\pm$0.4} & 11.9\,{\tableuncert$\pm$0.6} & 38.5\,{\tableuncert$\pm$0.2} & 15.3\,{\tableuncert$\pm$0.1} \\
+ adapter (ours) & \textbf{45.8}\,{\tableuncert$\pm$0.8} & \textbf{24.7}\,{\tableuncert$\pm$1.4} & \textbf{67.0}\,{\tableuncert$\pm$0.2} & \textbf{30.3}\,{\tableuncert$\pm$0.3} \\
base, no guidance & 45.1\,{\tableuncert$\pm$0.3} & 23.7\,{\tableuncert$\pm$0.1} & 65.9\,{\tableuncert$\pm$0.4} & 30.1\,{\tableuncert$\pm$0.1} \\
\midrule
\multicolumn{5}{@{}l}{\emph{SD-Turbo}} \\
distilled & 33.7\,{\tableuncert$\pm$0.9} & 14.0\,{\tableuncert$\pm$1.1} & 40.4\,{\tableuncert$\pm$0.8} & 16.8\,{\tableuncert$\pm$0.2} \\
+ adapter (ours) & \textbf{47.7}\,{\tableuncert$\pm$0.3} & \textbf{24.8}\,{\tableuncert$\pm$0.3} & \textbf{63.0}\,{\tableuncert$\pm$0.7} & \textbf{28.7}\,{\tableuncert$\pm$0.4} \\
SD 2.1, no guidance & 51.0\,{\tableuncert$\pm$0.7} & 28.6\,{\tableuncert$\pm$0.6} & 67.9\,{\tableuncert$\pm$0.3} & 33.3\,{\tableuncert$\pm$0.5} \\
\midrule
real, 500 per class & 50.3\,{\tableuncert$\pm$0.9} & 24.4\,{\tableuncert$\pm$1.0} & 70.1\,{\tableuncert$\pm$0.5} & 33.4\,{\tableuncert$\pm$0.5} \\
\bottomrule
\end{tabular}

  \caption{\textbf{Corrected images teach transferable representations.}
Linear-probe top-1 (\%) on real target data, using frozen ResNet-50 features learned on generated ImageNet-100
images; mean $\pm$ sd over three backbone seeds, one adapter run. IN-900: the 900 classes outside ImageNet-100.
Target supervision trains only the probe. Bold: adapted versus matched released generator.}
  \label{tab:transfer}
\end{minipage}\par\medskip
\paragraph{The learned representations transfer.} We freeze each ImageNet-100 classifier and fit a linear probe
on Food-101, Cars, Pets, and the 900 remaining ImageNet classes (\cref{tab:transfer}). Adapter-trained features gain
\nm{trGainMin}--\nm{trGainMax} points over features learned from the distilled releases. Klein's adapted images
produce features with accuracy comparable to its unguided base across all four targets, while requiring only two
denoiser evaluations per image. Only the probe sees target-dataset training images; its protocol is in
\cref{sec:supp-protocol}.

\subsection{Reference accuracies for other learners}
\label{sec:supp-learners}
The released-student and adapter comparisons are in \cref{tab:learners}.
\Cref{tab:supp-learners} supplies their unguided-base and real-data references without repeating those rows.
All five learners use matched generated sets; the adapter comparisons in \cref{tab:learners} use the first
trained adapter, as do the set metrics in \cref{tab:supp-quality}.

\par\medskip\noindent
\begin{minipage}{\linewidth}
  \captionsetup{type=table,hypcap=false}
  \centering
  \small
  \renewcommand{\arraystretch}{1.06}
\begin{tabular}{@{}lrrrrr@{}}
\toprule
training set & R50 & \shortstack{R50 +\\RandAugment} & \shortstack{Prior\\recipe} & ConvNeXt-T & ViT-S \\
\midrule
FLUX.2-klein base, no guidance & 45.1\,{\tableuncert$\pm$0.0} & 47.8\,{\tableuncert$\pm$0.2} & 41.1\,{\tableuncert$\pm$0.5} & 39.1\,{\tableuncert$\pm$0.8} & 25.4\,{\tableuncert$\pm$0.2} \\
SD 2.1, no guidance & 49.8\,{\tableuncert$\pm$0.7} & 52.8\,{\tableuncert$\pm$0.8} & 39.7\,{\tableuncert$\pm$0.3} & 42.4\,{\tableuncert$\pm$2.4} & 22.9\,{\tableuncert$\pm$0.2} \\
real images, 500 per class & 77.1\,{\tableuncert$\pm$0.2} & 80.3\,{\tableuncert$\pm$0.2} & 68.1\,{\tableuncert$\pm$0.9} & 72.3\,{\tableuncert$\pm$0.3} & 53.8\,{\tableuncert$\pm$0.4} \\
\bottomrule
\end{tabular}

  \caption{\textbf{Unguided-base and real-data references for the learner comparison.}
  ImageNet-100 top-1 (mean $\pm$ sd over three classifier seeds), using 500 images per class. Columns match the
  primary ResNet-50, ResNet-50 with RandAugment (2, 9), the reimplemented recipe of~\cite{adamkiewicz2026pretty},
  ConvNeXt-T (AdamW, RandAugment), and ViT-S/16. Released-student and adapter results under the four alternative
  learners are in \cref{tab:learners}; primary ResNet-50 results for the matched single-adapter sets are in
  \cref{tab:supp-quality}.}
  \label{tab:supp-learners}
\end{minipage}\par\medskip

\FloatBarrier\section{Prompt regimes, real-data availability, and shifted tests}

\subsection{Rich captions}

\label{sec:supp-captions}

\par\medskip\noindent
\begin{minipage}{\linewidth}
  \captionsetup{type=table,hypcap=false}
  \centering
  \tablebody
  \setlength{\tabcolsep}{3pt}
  \renewcommand{\arraystretch}{1.06}
\begin{tabular}{@{}lrrr@{}}
\toprule
sampler & templates & captions & LLM prompts \\
\midrule
SD 1.5, guidance 7.5 & 32.9\,{\tableuncert$\pm$0.5} & \textbf{42.2}\,{\tableuncert$\pm$0.7} & -- \\
SD 1.5, guidance 2 & 52.1\,{\tableuncert$\pm$0.4} & \textbf{57.3}\,{\tableuncert$\pm$0.8} & 50.2\,{\tableuncert$\pm$0.4} \\
\midrule
SD 2.1, guidance 7.5 & 30.0\,{\tableuncert$\pm$0.1} & \textbf{43.5}\,{\tableuncert$\pm$1.6} & -- \\
SD 2.1, guidance 2 & 49.5\,{\tableuncert$\pm$0.3} & \textbf{57.2}\,{\tableuncert$\pm$0.3} & 50.3\,{\tableuncert$\pm$0.1} \\
SD 2.1, no guidance & 49.8\,{\tableuncert$\pm$0.7} & \textbf{53.5}\,{\tableuncert$\pm$0.4} & -- \\
\midrule
SD-Turbo, 1 step & 15.8\,{\tableuncert$\pm$0.4} & \textbf{38.3}\,{\tableuncert$\pm$0.4} & 31.4\,{\tableuncert$\pm$0.5} \\
SD-Turbo, 4 steps & 18.3\,{\tableuncert$\pm$0.3} & \textbf{32.4}\,{\tableuncert$\pm$0.2} & 29.6\,{\tableuncert$\pm$0.7} \\
SD-Turbo, base-first & 36.4\,{\tableuncert$\pm$0.9} & \textbf{42.8}\,{\tableuncert$\pm$0.5} & 37.6\,{\tableuncert$\pm$0.6} \\
SD-Turbo + adapter & 43.3\,{\tableuncert$\pm$0.2} & \textbf{53.2}\,{\tableuncert$\pm$0.3} & 45.8\,{\tableuncert$\pm$0.5} \\
\midrule
SDXL, guidance 5 & 28.6\,{\tableuncert$\pm$0.3} & \textbf{42.7}\,{\tableuncert$\pm$0.5} & -- \\
SDXL-Turbo & 12.9\,{\tableuncert$\pm$0.5} & \textbf{40.5}\,{\tableuncert$\pm$1.0} & -- \\
\midrule
klein base, guidance 4 & 23.6\,{\tableuncert$\pm$0.5} & \textbf{55.1}\,{\tableuncert$\pm$1.7} & -- \\
klein base, guidance 2 & 34.8\,{\tableuncert$\pm$0.7} & \textbf{61.6}\,{\tableuncert$\pm$1.2} & -- \\
klein base, no guidance & 45.1\,{\tableuncert$\pm$0.0} & \textbf{63.0}\,{\tableuncert$\pm$1.3} & 42.9\,{\tableuncert$\pm$0.7} \\
FLUX.2-klein & 17.1\,{\tableuncert$\pm$0.4} & \textbf{55.0}\,{\tableuncert$\pm$0.8} & 37.9\,{\tableuncert$\pm$0.2} \\
FLUX.2-klein, base-first & 39.9\,{\tableuncert$\pm$0.3} & \textbf{60.8}\,{\tableuncert$\pm$0.2} & 43.1\,{\tableuncert$\pm$0.6} \\
FLUX.2-klein + adapter & 44.1\,{\tableuncert$\pm$0.5} & \textbf{61.5}\,{\tableuncert$\pm$0.4} & 39.6\,{\tableuncert$\pm$0.3} \\
\bottomrule
\end{tabular}

  \caption{\textbf{Real-image captions improve every evaluated sampler.} Top-1 on ImageNet-100 with the eight class-name templates, with captions
  of real training images, and with LLM-written class descriptions as prompts (one generated set per cell,
  mean $\pm$ SD over three classifier seeds; --:
  not run). Bold: best prompt source per sampler.}
  \label{tab:supp-captions}
\end{minipage}\par\medskip

Both rich-prompt regimes use Qwen2.5-VL-7B-Instruct~\cite{bai2025qwen25vl}. Real-image captions request
5--6 sentences describing image content and viewpoint; image-free descriptions request 3--4 sentences from
class names and seeded scene/viewpoint hints. Captions provide more specific content and layout information than class-name templates.
Using them as generation prompts raises every evaluated set, particularly the distilled models (\cref{tab:supp-captions}), and shrinks the gaps: with captions, guidance 7.5 against 2 costs SD~2.1
\gpa{capGuideSD} points (\gpa{guideSDTwo} with templates), distillation costs FLUX.2-klein \gpa{capDistKlein} points
against its unguided base (\gpa{distKleinUnguided}), and SDXL's distillation gap all but closes (\gp{capDistSDXL}
\gpci{capDistSDXL}, against \gp{distSDXL} with templates). The adapters (their first training runs) keep
\nm{capShareKlein}\% and \nm{capShareSD}\% of the unguided bases' gains under captions (\cref{tab:supp-captions}).
LLM-written descriptions raise the distilled models as well and leave FLUX.2-klein \gpa{llmGapKlein} points below its
unguided base; there the SD-Turbo adapter keeps \nm{llmShareSD}\% of its base's gain and klein's \nm{llmShareKlein}\%.
These single-set comparisons transfer the same template-trained adapter to different prompts. The replicated
LLM-description evaluation below separately tests the adaptation prompt source. Real-image-caption-specific
adapters are not evaluated; the COCO-caption adapters are a separate test of adaptation prompts.

\paragraph{Operating points under richer prompts.} The prompt-matched comparisons in
\cref{tab:supp-captions} retain both the adapter's benefit and the competitive alternatives.
For klein in the single-set comparison of \cref{tab:supp-captions}, base-first has the highest observed accuracy, while the template-trained adapter uses two
student evaluations and the released model avoids adapter preparation. SD-Turbo retains substantial adapter
gains under both rich-prompt regimes (\cref{tab:prompts}). These image-free descriptions use class names and
seeded scene/viewpoint hints; real-image captions require access to training images. Preparation and inference
requirements are compared separately in \cref{tab:supp-deployment}.

\subsection{Fitting adapters on image-free LLM descriptions}
\label{sec:supp-e50}
We fit one additional adapter per generator on 16,000 image-free descriptions, 160 per ImageNet-100 class.
Rank, endpoint, optimizer, training steps, and inference schedule match the template-trained adapter.
Each description is paired with one initial noise sample; the template targets instead use 20 noises for
each of eight templates. Klein uses the four-anchor teacher grid examined in the target ablations;
SD-Turbo's teacher grid is unchanged. Adaptation descriptions have no exact duplicates in the evaluation
pools, and their noise seeds are disjoint from generation seeds.

Three evaluation datasets use independently sampled descriptions and noise, with prompts and noise matched
across samplers within each dataset. Every generated dataset trains three classifier seeds. Each adapter
is a single training run. \Cref{tab:supp-e50-replication} separates within-dataset classifier SD from the SD
of dataset means. The latter still includes classifier noise. Pooled paired class-bootstrap intervals average
all nine runs first; they condition on the three generated datasets and fitted checkpoints, without covering
adapter-training variation or resampling dataset realizations.

\par\medskip\noindent
\begin{minipage}{\linewidth}
  \captionsetup{type=table,hypcap=false}
  \centering\small
  \setlength{\tabcolsep}{4pt}
\begin{tabular}{@{}lrrrrrr@{}}
\toprule
Sampler & Mean & Classifier SD & Dataset-mean SD & Set 1 & Set 2 & Set 3 \\
\midrule
\multicolumn{7}{@{}l}{\emph{FLUX.2-klein}} \\
Released & 37.1 & 0.26 & 0.71 & 37.9 & 36.5 & 37.0 \\
Base-first & 42.2 & 2.39 & 1.43 & 43.1 & 40.6 & 43.0 \\
Template-fitted & 39.2 & 0.55 & 0.61 & 39.6 & 39.5 & 38.5 \\
LLM-fitted & 42.5 & 0.69 & 0.70 & 43.1 & 42.7 & 41.7 \\
\midrule
\multicolumn{7}{@{}l}{\emph{SD-Turbo}} \\
Released & 29.2 & 0.71 & 0.38 & 29.6 & 29.1 & 28.9 \\
Base-first & 37.9 & 0.68 & 0.26 & 37.6 & 38.1 & 37.9 \\
Template-fitted & 45.9 & 0.60 & 0.27 & 45.8 & 45.7 & 46.2 \\
LLM-fitted & 44.9 & 0.39 & 0.22 & 44.9 & 45.0 & 44.6 \\
\bottomrule
\end{tabular}

  \caption{\textbf{Independent prompt and noise draws test the repeatability of the comparisons.}
  ImageNet-100 ResNet-50 top-1 (\%), 500 generated images/class, LLM generation prompts. Mean: nine classifier
  runs across three generated datasets. Classifier SD pools within-dataset variation; dataset-mean SD uses
  the three means and still contains classifier variation. Set columns give means of three classifier seeds.
  Adapted arms use one training run; every run is retained.}
  \label{tab:supp-e50-replication}
\end{minipage}\par\medskip

\par\medskip\noindent
\begin{minipage}{\linewidth}
  \captionsetup{type=table,hypcap=false}
  \centering\small
  \setlength{\tabcolsep}{4pt}
\begin{tabular}{@{}lrr@{}}
\toprule
Adapter comparison & FLUX.2-klein & SD-Turbo \\
\midrule
\multicolumn{3}{@{}l}{\emph{LLM generation prompts: three datasets, three classifier seeds each}} \\
LLM-fitted $-$ template-fitted & \textbf{+3.3}\,{\tableuncert[+2.4, +4.2]} & $-$1.1\,{\tableuncert[$-$1.6, $-$0.5]} \\
LLM-fitted $-$ released & \textbf{+5.4}\,{\tableuncert[+3.4, +7.3]} & \textbf{+15.6}\,{\tableuncert[+12.9, +18.5]} \\
LLM-fitted $-$ base-first & +0.3\,{\tableuncert[$-$0.7, +1.3]} & \textbf{+7.0}\,{\tableuncert[+5.6, +8.3]} \\
Template-fitted $-$ released & \textbf{+2.1}\,{\tableuncert[+0.2, +4.0]} & \textbf{+16.7}\,{\tableuncert[+14.1, +19.4]} \\
Template-fitted $-$ base-first & $-$3.0\,{\tableuncert[$-$4.2, $-$1.9]} & \textbf{+8.0}\,{\tableuncert[+6.6, +9.5]} \\
\midrule
\multicolumn{3}{@{}l}{\emph{Template generation prompts: one dataset, three classifier seeds}} \\
LLM-fitted $-$ template-fitted & $-$1.4\,{\tableuncert[$-$2.8, $-$0.1]} & $-$1.1\,{\tableuncert[$-$2.0, $-$0.2]} \\
\bottomrule
\end{tabular}

  \caption{\textbf{The two generators favor different fitting prompts on ImageNet-100.}
  Accuracy differences in points [paired 95\% class-bootstrap CI]. Upper block averages three generated
  datasets and three classifier seeds each; lower block uses template generation prompts and one dataset.
  Each adapter uses one training run. The first row compares LLM-fitted with template-fitted adapters.
  Bold: interval above zero.}
  \label{tab:supp-e50-fitting}
\end{minipage}\par\medskip

The LLM-fitted klein adapter improves on the template-trained adapter on every dataset, but has no resolved
advantage over base-first (\cref{tab:supp-e50-fitting}). The template-trained SD-Turbo adapter remains better
under both prompt styles. The preferred fitted adapter therefore depends on the generator and generation
prompts in these comparisons.

\paragraph{Classifier-run sensitivity.} One klein base-first classifier trains poorly on the second dataset,
ending at \nm{e50PoorRun}\% versus \nm{e50OtherRunOne}\% and \nm{e50OtherRunTwo}\% for the other seeds on the same images. The primary analysis retains
it because no exclusion rule was specified. A post-hoc sensitivity analysis omitting that run changes the
LLM-fitted-adapter minus base-first gap from \nm{e50LlmBaseFirstKlein} (\nm{e50LlmBaseFirstKleinCI}) to
\nm{e50Sensitivity} (\nm{e50SensitivityCI}) points; both leave the
advantage unresolved. The template-trained adapter remains below base-first. This sensitivity is not used
to select an adapter.

\subsection{Frozen-adapter evaluation with LLM prompts on ImageNet-1k}
For each generator, we select the adapter with higher pooled ImageNet-100 accuracy under LLM prompts:
the LLM-fitted klein adapter and the template-trained SD-Turbo adapter. The selection rule is specified
before the comparisons, and the choices are recorded before any ImageNet-1k classifier training. Endpoints,
ranks, and sampling schedules remain fixed. The scale evaluation uses 100 images for each of 1,000 classes,
one generated dataset per sampler, and three classifier seeds; it does not evaluate the 1,281-images/class regime.

\par\medskip\noindent
\begin{minipage}{\linewidth}
  \captionsetup{type=table,hypcap=false}
  \centering\small
  \setlength{\tabcolsep}{4pt}
\begin{tabular}{@{}llrrrr@{}}
\toprule
Sampler & Generation prompts & NFE & Top-1 & vs. LLM released & vs. LLM base-first \\
\midrule
\multicolumn{6}{@{}l}{\emph{FLUX.2-klein}} \\
Released & LLM & 4 & 16.0\,{\tableuncert$\pm$0.4} & -- & -- \\
Base-first & LLM & 4 & 16.9\,{\tableuncert$\pm$0.2} & \textbf{+0.9}\,{\tableuncert[+0.4, +1.3]} & -- \\
Original adapter & Templates & 2 & 18.6\,{\tableuncert$\pm$0.0} & \textbf{+2.7}\,{\tableuncert[+2.0, +3.4]} & \textbf{+1.8}\,{\tableuncert[+1.2, +2.4]} \\
LLM-fitted adapter & LLM & 2 & 14.5\,{\tableuncert$\pm$0.2} & $-$1.5\,{\tableuncert[$-$2.0, $-$1.0]} & $-$2.4\,{\tableuncert[$-$2.8, $-$2.0]} \\
Template fit (post hoc) & LLM & 2 & 12.6\,{\tableuncert$\pm$0.4} & $-$3.3\,{\tableuncert[$-$3.9, $-$2.8]} & $-$4.2\,{\tableuncert[$-$4.7, $-$3.8]} \\
Unguided klein base & LLM & 20 & 13.5\,{\tableuncert$\pm$0.1} & $-$2.4\,{\tableuncert[$-$3.0, $-$1.9]} & $-$3.3\,{\tableuncert[$-$3.8, $-$2.9]} \\
\midrule
\multicolumn{6}{@{}l}{\emph{SD-Turbo}} \\
Released & LLM & 4 & 12.0\,{\tableuncert$\pm$0.1} & -- & -- \\
Base-first & LLM & 4 & 13.7\,{\tableuncert$\pm$0.4} & \textbf{+1.7}\,{\tableuncert[+1.1, +2.4]} & -- \\
Original adapter & Templates & 3 & 19.1\,{\tableuncert$\pm$0.1} & \textbf{+7.1}\,{\tableuncert[+6.3, +7.9]} & \textbf{+5.4}\,{\tableuncert[+4.8, +6.0]} \\
Template-fitted adapter & LLM & 3 & 18.9\,{\tableuncert$\pm$0.4} & \textbf{+6.9}\,{\tableuncert[+6.2, +7.7]} & \textbf{+5.2}\,{\tableuncert[+4.8, +5.6]} \\
Unguided SD 2.1 & LLM & 30 & 13.5\,{\tableuncert$\pm$0.1} & \textbf{+1.5}\,{\tableuncert[+0.8, +2.3]} & $-$0.2\,{\tableuncert[$-$0.8, +0.3]} \\
\bottomrule
\end{tabular}

  \caption{\textbf{The original template-based adapters remain useful against richer-prompt alternatives.}
  ImageNet-1k ResNet-50 top-1 (\%, mean $\pm$ SD over three classifier seeds on one generated dataset),
  100 images/class. NFE counts denoiser evaluations. Differences in points [paired 95\% class-bootstrap CI]
  compare each procedure with the released or base-first sampler using LLM prompts. Template rows change
  both adaptation and generation prompts relative to these references. The primary LLM rows use the fitting
  source selected on ImageNet-100 and frozen before this evaluation. Klein's additional template-fit row
  is post hoc. Unguided-base rows are follow-up comparisons added after the adapter evaluation.
  Every adapter uses one training run.
  Bold: interval above zero.}
  \label{tab:supp-e50-in1k}
\end{minipage}\par\medskip

The original template-based adapters exceed the released and base-first samplers using LLM descriptions
for both generators (\cref{tab:supp-e50-in1k}). These end-to-end comparisons change the generation prompts
as well as the sampler. Under matched LLM generation prompts, the selected SD-Turbo adapter improves on
both references, while the selected klein adapter falls below both. On the 900 classes outside adapter
fitting, their gains over the released students are \nm{e50NineHundredkd_llm} (\nm{e50NineHundredkd_llmCI})
for klein and \nm{e50NineHundredsdt_ad1} (\nm{e50NineHundredsdt_ad1CI}) for SD-Turbo, retaining 1,000-way predictions.
Those classes appeared in the earlier template-prompt development pilot.

\paragraph{Post-hoc template-adapter check.} We additionally evaluate klein's original template-trained
adapter under the same ImageNet-1k LLM prompts, seeds, and classifier protocol. This comparison is specified
after inspecting the primary scale results. It gives \nm{e50PostHocKleinVsReleased} points
(\nm{e50PostHocKleinVsReleasedCI}) against the released student and \nm{e50PostHocKleinVsBaseFirst}
(\nm{e50PostHocKleinVsBaseFirstCI}) against base-first. On the 900 classes outside adapter fitting, its
gain over the released student is \nm{e50PostHocKleinNineHundred} (\nm{e50PostHocKleinNineHundredCI}).
The LLM-fitted adapter remains better than the template-trained one by \nm{e50LlmVsTemplateAtScale} points
(\nm{e50LlmVsTemplateAtScaleCI}), so the development rule selects the better of the two tested adapters.
Nevertheless, neither improves on the released student under these generation prompts. This identifies a
limit for the tested klein adapters at 100 images per class; it does not explain its cause or establish the outcome at
1,281 images per class.

\paragraph{Follow-up teacher and data-size checks.} The unguided-base comparisons use the same ImageNet-1k
LLM prompts, image seeds, and classifier protocol as the adapter evaluation (\cref{tab:supp-e50-in1k}).
These full-base samplers differ from the adapters' fine-grid intermediate targets followed by student
completion, so the comparisons do not isolate the cause of klein's loss.

On ImageNet-100 with LLM generation prompts, the LLM-fitted klein adapter's gap to the released student
changes from \gp{e52KleinLlm100} points \gpci{e52KleinLlm100} at 100 images per class to
\gp{e52KleinLlm500} \gpci{e52KleinLlm500} at 500. This comparison uses one generated dataset per sampler and
its 100-image-per-class subset, three classifier seeds, and the same fitted adapter. Both sizes use 100
training epochs, so the larger sets also receive more optimizer updates. The reversal shows that the
failure does not require 1,000 classes, but does not separate the effects of dataset size and optimization budget.

\subsection{Augmenting real training data}

\label{sec:supp-mixtures}

\paragraph{Benefits with limited real data.} At 500 real images per class, adding adapted images improves
accuracy over both real-only training and augmentation with released-student images. Released klein images
provide no resolved improvement over the matched real-only reference, whereas base-first images improve it.
With only 50 real images per class, the real-only reference reaches \tp{low_real50}\%. The adapter's
advantage over released-student images grows to \gp{lowAdKlein} points for klein and \gp{lowAdSD} for
SD-Turbo, and adapted images also outperform unguided-base images. Paired differences and uncertainty
are collected in \cref{tab:supp-contrasts}.

\paragraph{At full real-data scale.} Adding adapted images to the full real ImageNet-1k training set improves
ImageNet-Sketch and ImageNet-R accuracy while slightly lowering in-distribution accuracy. Adding released
klein images instead lowers all three accuracies. Adapted images therefore provide the better synthetic
augmentation in this comparison, with the benefit over real-only training concentrated on the shifted
test sets (\cref{tab:supp-ood,tab:supp-contrasts}).

\subsection{Accuracy on shifted test sets}

\label{sec:supp-ood}

\par\medskip\noindent
\begin{minipage}{\linewidth}
  \captionsetup{type=table,hypcap=false}
  \centering
  \small
  \renewcommand{\arraystretch}{1.06}
\begin{tabular}{@{}lrrrrr@{}}
\toprule
training set & in distribution & V2 & Sketch & R & A \\
\midrule
\multicolumn{6}{@{}l}{\emph{ImageNet-100, generated only}} \\
real images, 500 per class & 77.1\,{\tableuncert$\pm$0.2} & 66.1\,{\tableuncert$\pm$1.4} & 21.1\,{\tableuncert$\pm$1.1} & 30.8\,{\tableuncert$\pm$1.1} & 4.3\,{\tableuncert$\pm$0.7} \\
FLUX.2-klein, distilled & 17.1\,{\tableuncert$\pm$0.4} & 15.4\,{\tableuncert$\pm$0.8} & 5.0\,{\tableuncert$\pm$0.1} & 6.8\,{\tableuncert$\pm$0.5} & 0.5\,{\tableuncert$\pm$0.2} \\
\quad base-first & 39.9\,{\tableuncert$\pm$0.3} & 34.9\,{\tableuncert$\pm$0.2} & 16.3\,{\tableuncert$\pm$0.4} & 22.6\,{\tableuncert$\pm$0.3} & 2.8\,{\tableuncert$\pm$0.1} \\
\quad + adapter (ours) & 43.8\,{\tableuncert$\pm$0.4} & 37.4\,{\tableuncert$\pm$0.9} & 22.5\,{\tableuncert$\pm$0.7} & 27.9\,{\tableuncert$\pm$0.8} & 3.6\,{\tableuncert$\pm$0.4} \\
\quad base, no guidance & 45.1\,{\tableuncert$\pm$0.0} & 37.4\,{\tableuncert$\pm$1.0} & 23.0\,{\tableuncert$\pm$0.7} & 30.3\,{\tableuncert$\pm$1.8} & 3.7\,{\tableuncert$\pm$0.4} \\
SD-Turbo, distilled & 18.3\,{\tableuncert$\pm$0.3} & 14.9\,{\tableuncert$\pm$0.4} & 4.9\,{\tableuncert$\pm$0.4} & 4.6\,{\tableuncert$\pm$0.5} & 2.2\,{\tableuncert$\pm$0.5} \\
\quad base-first & 36.4\,{\tableuncert$\pm$0.9} & 28.8\,{\tableuncert$\pm$1.4} & 11.1\,{\tableuncert$\pm$0.6} & 13.7\,{\tableuncert$\pm$0.1} & 3.8\,{\tableuncert$\pm$0.4} \\
\quad + adapter (ours) & 43.7\,{\tableuncert$\pm$0.6} & 36.8\,{\tableuncert$\pm$1.4} & 12.7\,{\tableuncert$\pm$1.4} & 16.4\,{\tableuncert$\pm$0.4} & 2.6\,{\tableuncert$\pm$0.4} \\
\quad SD 2.1, no guidance & 49.8\,{\tableuncert$\pm$0.7} & 41.0\,{\tableuncert$\pm$0.6} & 18.6\,{\tableuncert$\pm$1.8} & 26.4\,{\tableuncert$\pm$0.7} & 2.8\,{\tableuncert$\pm$0.3} \\
Z-Image-Turbo, distilled & 14.0\,{\tableuncert$\pm$0.0} & 12.0\,{\tableuncert$\pm$0.3} & 5.7\,{\tableuncert$\pm$0.4} & 6.0\,{\tableuncert$\pm$0.1} & 0.9\,{\tableuncert$\pm$0.1} \\
\quad + adapter (ours) & 35.1\,{\tableuncert$\pm$0.5} & 30.0\,{\tableuncert$\pm$1.2} & 12.9\,{\tableuncert$\pm$0.6} & 13.2\,{\tableuncert$\pm$0.6} & 2.1\,{\tableuncert$\pm$0.5} \\
\midrule
\multicolumn{6}{@{}l}{\emph{ImageNet-100, 500 real + 500 generated}} \\
real only, trained as long & 78.8\,{\tableuncert$\pm$0.0} & 67.0\,{\tableuncert$\pm$0.9} & 21.6\,{\tableuncert$\pm$0.4} & 31.0\,{\tableuncert$\pm$0.9} & 3.8\,{\tableuncert$\pm$0.7} \\
+ FLUX.2-klein, distilled & 78.7\,{\tableuncert$\pm$0.4} & 67.8\,{\tableuncert$\pm$1.6} & 23.7\,{\tableuncert$\pm$0.6} & 33.3\,{\tableuncert$\pm$0.9} & 5.1\,{\tableuncert$\pm$0.3} \\
+ FLUX.2-klein adapter & 80.4\,{\tableuncert$\pm$0.1} & 70.1\,{\tableuncert$\pm$1.6} & 31.8\,{\tableuncert$\pm$0.8} & 38.6\,{\tableuncert$\pm$1.3} & 5.1\,{\tableuncert$\pm$0.5} \\
+ SD-Turbo, distilled & 79.3\,{\tableuncert$\pm$0.2} & 69.0\,{\tableuncert$\pm$0.8} & 25.9\,{\tableuncert$\pm$0.9} & 33.9\,{\tableuncert$\pm$0.9} & 5.1\,{\tableuncert$\pm$0.5} \\
+ SD-Turbo adapter & 82.2\,{\tableuncert$\pm$0.3} & 70.1\,{\tableuncert$\pm$0.9} & 29.0\,{\tableuncert$\pm$0.3} & 36.9\,{\tableuncert$\pm$0.5} & 4.9\,{\tableuncert$\pm$0.4} \\
\midrule
\multicolumn{6}{@{}l}{\emph{ImageNet-1k, 1,281 per class}} \\
FLUX.2-klein, distilled & 8.4\,{\tableuncert$\pm$0.1} & 7.1\,{\tableuncert$\pm$0.1} & 2.2\,{\tableuncert$\pm$0.1} & 1.8\,{\tableuncert$\pm$0.1} & 0.1\,{\tableuncert$\pm$0.0} \\
\quad + adapter (ours) & 34.2\,{\tableuncert$\pm$0.1} & 29.0\,{\tableuncert$\pm$0.1} & 13.6\,{\tableuncert$\pm$0.1} & 15.0\,{\tableuncert$\pm$0.0} & 1.3\,{\tableuncert$\pm$0.1} \\
SD-Turbo, distilled & 12.7\,{\tableuncert$\pm$0.2} & 10.5\,{\tableuncert$\pm$0.1} & 2.9\,{\tableuncert$\pm$0.2} & 4.2\,{\tableuncert$\pm$0.0} & 0.5\,{\tableuncert$\pm$0.1} \\
\quad + adapter (ours) & 32.1\,{\tableuncert$\pm$0.3} & 27.1\,{\tableuncert$\pm$0.4} & 7.4\,{\tableuncert$\pm$0.1} & 10.1\,{\tableuncert$\pm$0.1} & 1.2\,{\tableuncert$\pm$0.1} \\
\midrule
\multicolumn{6}{@{}l}{\emph{ImageNet-1k, all 1.28M real + 1.28M generated}} \\
real only, trained as long & 76.7\,{\tableuncert$\pm$0.2} & 65.2\,{\tableuncert$\pm$0.3} & 24.4\,{\tableuncert$\pm$0.1} & 25.4\,{\tableuncert$\pm$0.2} & 2.5\,{\tableuncert$\pm$0.1} \\
+ FLUX.2-klein, distilled & 76.0\,{\tableuncert$\pm$0.0} & 64.5\,{\tableuncert$\pm$0.2} & 24.0\,{\tableuncert$\pm$0.5} & 24.8\,{\tableuncert$\pm$0.5} & 1.7\,{\tableuncert$\pm$0.1} \\
+ FLUX.2-klein adapter & 76.5\,{\tableuncert$\pm$0.2} & 65.0\,{\tableuncert$\pm$0.2} & 27.7\,{\tableuncert$\pm$0.6} & 27.9\,{\tableuncert$\pm$0.4} & 2.6\,{\tableuncert$\pm$0.1} \\
\bottomrule
\end{tabular}

  \caption{\textbf{Classifier accuracy on in-distribution and shifted test sets.} Top-1 (\%, mean $\pm$ sd over classifier runs;
  the adapters on ImageNet-100 pool three training runs of three seeds) of the ResNet-50 classifiers of the main text on
  their own validation images (in distribution) and on ImageNet-V2, -Sketch, -R and -A restricted to their classes,
  with the argmax over all of the classifier's classes. On ImageNet-100, V2 and Sketch cover all 100 classes, R 19 and
  A 15; on ImageNet-1k, V2 and Sketch cover all 1,000, R and A 200 each. Last block: all 1,281,167 real ImageNet-1k
  training images alone for 200 epochs, or with 1.28M generated images for 100 epochs (the same number of updates).}
  \label{tab:supp-ood}
\end{minipage}\par\medskip

We evaluate the saved classifiers on ImageNet-V2~\cite{recht2019imagenet}, -Sketch~\cite{wang2019learning}, -R~\cite{hendrycks2021many}, and -A~\cite{hendrycks2021natural} without retraining (\cref{tab:supp-ood}). The
evaluation reproduces each classifier's validation accuracy before its numbers are kept. Gaps in the main text are
paired differences of per-class top-1 with 95\% bootstrap intervals over the classes a set covers.

\FloatBarrier\section{Diagnostic metrics and their scope}
\label{sec:mechanism}
\label{sec:supp-metrics}
We compare the fixed collection of \nm{eNSets} template-prompt sets with real images in feature space.
These diagnostics support \cref{sec:discussion}; they neither determine the adapter objective nor establish
causal mediation. Later evaluations do not expand this collection.

\subsection{Metric comparisons and collection sensitivity}

\par\medskip\noindent
\begin{minipage}{\linewidth}
  \captionsetup{type=table,hypcap=false}
  \centering
  \tablebody
  \setlength{\tabcolsep}{4pt}
  \renewcommand{\arraystretch}{1.06}
\begin{tabular}{@{}lrrrrrrr@{}}
\toprule
metric & within families & pooled & FLUX.2-klein & SD 2.1 & Z-Image & SDXL & held-out family \\
\midrule
recall, DINOv3 & +0.87 {\tableuncert[+0.81, +0.93]} & +0.91 & +0.95 & +0.91 & +0.97 & +0.20 & +0.85 \\
recall, CLIP & +0.86 {\tableuncert[+0.79, +0.91]} & +0.89 & +0.94 & +0.86 & +0.98 & +0.24 & +0.83 \\
Fr\'echet distance, CLIP & $-$0.86 {\tableuncert[$-$0.91, $-$0.79]} & $-$0.88 & $-$0.92 & $-$0.85 & $-$0.98 & $-$0.44 & +0.84 \\
Fr\'echet distance, DINOv3 & $-$0.74 {\tableuncert[$-$0.82, $-$0.63]} & $-$0.72 & $-$0.90 & $-$0.60 & $-$0.98 & $-$0.05 & +0.62 \\
coverage, DINOv3 & +0.30 {\tableuncert[+0.15, +0.45]} & +0.03 & +0.34 & +0.22 & +0.76 & +0.21 & $-$0.28 \\
coverage, CLIP & +0.63 {\tableuncert[+0.52, +0.72]} & +0.60 & +0.75 & +0.51 & +0.82 & +0.22 & +0.45 \\
precision, DINOv3 & $-$0.57 {\tableuncert[$-$0.69, $-$0.42]} & $-$0.58 & $-$0.61 & $-$0.47 & $-$0.97 & $-$0.26 & +0.49 \\
density, DINOv3 & $-$0.59 {\tableuncert[$-$0.70, $-$0.45]} & $-$0.57 & $-$0.64 & $-$0.49 & $-$0.93 & $-$0.23 & +0.44 \\
Vendi, DINOv3 & +0.78 {\tableuncert[+0.68, +0.86]} & +0.82 & +0.83 & +0.78 & +0.99 & +0.21 & +0.78 \\
Vendi, CLIP & +0.76 {\tableuncert[+0.65, +0.83]} & +0.79 & +0.84 & +0.68 & +0.98 & +0.28 & +0.73 \\
FID (Inception) & $-$0.66 {\tableuncert[$-$0.76, $-$0.55]} & $-$0.60 & $-$0.83 & $-$0.38 & $-$0.91 & $-$0.54 & +0.47 \\
Inception Score & $-$0.34 {\tableuncert[$-$0.48, $-$0.17]} & $-$0.48 & $-$0.44 & $-$0.25 & $-$0.36 & +0.17 & +0.38 \\
recognizability & $-$0.54 {\tableuncert[$-$0.66, $-$0.39]} & $-$0.43 & $-$0.65 & $-$0.41 & $-$0.87 & $-$0.01 & +0.21 \\
frozen DINOv3, linear probe & +0.69 {\tableuncert[+0.58, +0.76]} & +0.41 & +0.65 & +0.77 & +0.73 & +0.59 & +0.13 \\
frozen DINOv3, $k$-NN & +0.33 {\tableuncert[+0.18, +0.46]} & +0.21 & +0.42 & +0.04 & +0.73 & +0.63 & +0.02 \\
contrast & $-$0.49 {\tableuncert[$-$0.59, $-$0.36]} & $-$0.40 & $-$0.48 & $-$0.63 & $-$0.18 & $-$0.14 & +0.24 \\
sharpness & $-$0.30 {\tableuncert[$-$0.42, $-$0.16]} & +0.01 & $-$0.40 & $-$0.49 & +0.74 & +0.48 & $-$0.42 \\
saturation & $-$0.32 {\tableuncert[$-$0.43, $-$0.18]} & $-$0.37 & $-$0.36 & $-$0.36 & +0.50 & $-$0.35 & +0.31 \\
colorfulness & $-$0.41 {\tableuncert[$-$0.52, $-$0.28]} & $-$0.36 & $-$0.64 & $-$0.23 & +0.15 & $-$0.24 & +0.27 \\
high-frequency share & +0.01 {\tableuncert[$-$0.15, +0.15]} & +0.25 & +0.06 & $-$0.41 & +0.66 & +0.66 & +0.27 \\
spectral slope & +0.08 {\tableuncert[$-$0.08, +0.23]} & +0.27 & +0.18 & $-$0.24 & +0.67 & +0.42 & +0.32 \\
aesthetic score & $-$0.46 {\tableuncert[$-$0.58, $-$0.31]} & $-$0.32 & $-$0.56 & $-$0.35 & $-$0.90 & +0.26 & +0.09 \\
PickScore & $-$0.55 {\tableuncert[$-$0.69, $-$0.39]} & $-$0.59 & $-$0.67 & $-$0.43 & $-$0.49 & $-$0.21 & +0.58 \\
\midrule
aesthetic score given recall & +0.26 {\tableuncert[+0.08, +0.39]} & & & & & & \\
PickScore given recall & +0.11 {\tableuncert[$-$0.10, +0.30]} & & & & & & \\
\bottomrule
\end{tabular}

  \caption{\textbf{Set metrics against training value.} Spearman correlations with ResNet-50 top-1 on the
  \nm{eNSets} template-prompt ImageNet-100 sets: family-centered ranks [95\% configuration-bootstrap CI],
  pooled ranks, and within-family ranks (at least eight sets). Distances favor negative correlations.
  Held-out family: correlation of out-of-fold monotone predictions with top-1, pooled across families;
  positive values favor either metric direction. Bottom: partial correlations given DINOv3 recall.}
  \label{tab:supp-metrics}
\end{minipage}\par\medskip

DINOv3 recall, CLIP recall, and CLIP Fr\'echet distance have similar aggregate associations with accuracy,
with overlapping family-centered intervals (\cref{tab:supp-metrics}). Coverage, Vendi, and FID are weaker;
recognizability, Inception Score, and preference scores correlate negatively with accuracy.

\paragraph{Ranking protocol and limits.} Within family $f$, we transform metric and accuracy ranks to
$(\operatorname{rank}_f-(n_f+1)/2)/n_f$ and correlate the concatenated ranks. Intervals use 10,000 within-family
bootstrap resamples of sets (seed zero). For held-out-family prediction, we fit isotonic regression on the
other families, choosing its direction from their Spearman sign. Tied inputs are averaged with multiplicity
weights; predictions interpolate linearly between knots and clip at endpoints. We refit in each bootstrap
resample and correlate held-out predictions with accuracy within families and pooled across families
(\cref{tab:supp-heldout}). This tests ranking, not calibration; SDXL and FLUX.1-dev remain uncertain.

The bootstrap retains unequal family sizes and does not account for shared prompts, image seeds, neighboring
sampling plans, or the choice of configuration collection. The intervals condition on this collection;
held-out-family checks assess transfer only across the studied families.

\par\medskip\noindent
\begin{minipage}{\linewidth}
  \captionsetup{type=table,hypcap=false}
  \centering
  \small
  \setlength{\tabcolsep}{4pt}
\begin{tabular}{@{}lrr@{}}
\toprule
Held-out family & sets & $\rho$ [95\% CI] \\
\midrule
FLUX.2-klein & 89 & +0.94 {\tableuncert[+0.90, +0.97]} \\
SD 2.1 / SD-Turbo & 64 & +0.90 {\tableuncert[+0.82, +0.94]} \\
Z-Image & 13 & +0.97 {\tableuncert[+0.84, +1.00]} \\
SDXL & 12 & +0.21 {\tableuncert[$-$0.68, +0.79]} \\
FLUX.1-dev & 6 & +0.43 {\tableuncert[$-$0.78, +1.00]} \\
\midrule
Pooled predictions & 186 & +0.85 {\tableuncert[+0.79, +0.88]} \\
\bottomrule
\end{tabular}

  \caption{\textbf{Recall's ranking precision varies across held-out families.}
  Spearman correlation of predicted and observed ImageNet-100 top-1 [95\% family-stratified set-bootstrap CI].
  Each isotonic fit excludes the held-out family's outcomes. Rows require five sets; SD~1.5 has two and
  contributes only to the pooled evaluation.}
  \label{tab:supp-heldout}
\end{minipage}\par\medskip

\paragraph{Equal-family sensitivity.} We compare family-centered ranks with the mean within-family Spearman
correlation on the same 178 sets from klein, SD~2.1, SDXL, and Z-Image (\cref{tab:supp-family-weighting}).
Distances are negated; both summaries use the same 10,000 paired within-family bootstrap resamples.
CLIP Fr\'echet distance has the largest equal-family point estimate, but its paired difference from DINOv3
recall, \nm{equalFamilyFdMinusRecall} (95\% CI \nm{equalFamilyFdMinusRecallCI}), does not establish an ordering.
Equal weighting retains the configuration-dependence limitations above.

\par\medskip\noindent
\begin{minipage}{\linewidth}
  \captionsetup{type=table,hypcap=false}
  \centering
  \tablebody
  \setlength{\tabcolsep}{3pt}
\begin{tabular}{@{}lrr@{}}
\toprule
Metric & Family-centered & Equal-family \\
\midrule
DINOv3 recall & \textbf{+0.89}\,{\tableuncert[+0.83, +0.93]} & +0.76\,{\tableuncert[+0.55, +0.90]} \\
CLIP recall & +0.87\,{\tableuncert[+0.80, +0.91]} & +0.75\,{\tableuncert[+0.54, +0.89]} \\
$-$CLIP Fr\'echet & +0.87\,{\tableuncert[+0.80, +0.92]} & \textbf{+0.80}\,{\tableuncert[+0.61, +0.91]} \\
$-$FID & +0.66\,{\tableuncert[+0.54, +0.75]} & +0.67\,{\tableuncert[+0.46, +0.80]} \\
\bottomrule
\end{tabular}

  \caption{\textbf{Recall and CLIP Fr\'echet distance remain useful under equal family weight.}
  Ranking summaries [95\% configuration-bootstrap CI] on 178 sets from four families with at least eight
  sets. Left: concatenated centered ranks; right: equally weighted within-family Spearman correlations.
  Distance signs are reversed. Bold: largest point estimate. SD~1.5 (two sets) and FLUX.1-dev (six) remain
  in the original 186-set analysis.}
  \label{tab:supp-family-weighting}
\end{minipage}\par\medskip

\subsection{Adapted-set diagnostics}
\label{sec:supp-quality}

\par\medskip\noindent
\begin{minipage}{\linewidth}
  \captionsetup{type=table,hypcap=false}
  \centering
  \small
  \renewcommand{\arraystretch}{1.06}
\begin{tabular}{@{}lrrrrrr@{}}
\toprule
sampler & top-1 & FID & aesthetic & PickScore & recognized & recall \\
\midrule
\multicolumn{7}{@{}l}{\emph{FLUX.2-klein}} \\
distilled & 17.1\,{\tableuncert$\pm$0.4} & 47.6 & 5.28 & 22.06 & 0.83 & 0.009 \\
\quad + adapter (ours) & 44.1\,{\tableuncert$\pm$0.5} & 34.8 & 4.80 & 20.25 & 0.66 & 0.275 \\
base, no guidance & 45.1\,{\tableuncert$\pm$0.0} & 33.0 & 4.81 & 20.12 & 0.61 & 0.492 \\
base, default guidance & 23.6\,{\tableuncert$\pm$0.5} & 42.6 & 5.28 & 21.79 & 0.79 & 0.037 \\
\midrule
\multicolumn{7}{@{}l}{\emph{SD-Turbo}} \\
distilled & 18.3\,{\tableuncert$\pm$0.3} & 37.5 & 5.35 & 21.62 & 0.88 & 0.017 \\
\quad + adapter (ours) & 43.3\,{\tableuncert$\pm$0.2} & 30.2 & 5.21 & 20.94 & 0.80 & 0.164 \\
SD 2.1, no guidance & 49.8\,{\tableuncert$\pm$0.7} & 50.2 & 5.01 & 19.72 & 0.60 & 0.331 \\
SD 2.1, default guidance & 30.0\,{\tableuncert$\pm$0.1} & 28.0 & 5.44 & 21.46 & 0.91 & 0.039 \\
\midrule
\multicolumn{7}{@{}l}{\emph{Z-Image-Turbo}} \\
distilled & 14.0\,{\tableuncert$\pm$0.0} & 56.2 & 5.28 & 19.66 & 0.89 & 0.003 \\
\quad + adapter (ours) & 35.2\,{\tableuncert$\pm$0.6} & 24.9 & 5.19 & 21.54 & 0.87 & 0.081 \\
Z-Image, no guidance & 45.6\,{\tableuncert$\pm$0.7} & 29.4 & 4.92 & 18.56 & 0.69 & 0.386 \\
Z-Image, default guidance & 17.6\,{\tableuncert$\pm$0.3} & 34.6 & 5.34 & 19.32 & 0.89 & 0.033 \\
\bottomrule
\end{tabular}

  \caption{\textbf{Training-value gains do not require higher preference scores.} ImageNet-100 top-1
  (\%, mean $\pm$ sd over three classifier seeds) and metrics for sets of 500 images per class.
  FID and DINOv3 recall use real training images; lower FID is better. LAION aesthetic score and PickScore
  favor higher values. Recognized: fraction assigned to the intended class among the 100 by a pretrained
  ImageNet ResNet-50. Adapters use their first training runs; \cref{tab:supp-main-complete} pools three.}
  \label{tab:supp-quality}
\end{minipage}\par\medskip

Adaptation improves FID in \nm{qFidBetter} families while reducing recognizability in \nm{qRecogLower}
(\cref{tab:supp-quality}).

\paragraph{Class-level limits.}
\label{sec:supp-perclass}
For the first fitted klein and SD-Turbo adapters, classwise recall changes correlate inconsistently with
accuracy gains (\nm{pcRecallChangeKlein} and \nm{pcRecallChangeSD}, respectively). The set-level association
does not provide a reliable class-level explanation.

\section{Supporting accuracy contrasts}
\label{sec:supp-contrasts}

\par\medskip\noindent
\begin{minipage}{\linewidth}
  \captionsetup{type=table,hypcap=false}
  \centering
  \tablebody
  \setlength{\tabcolsep}{3pt}
\begin{tabular}{@{}lrr@{}}
\toprule
Comparison & Gain & 95\% CI \\
\midrule
\multicolumn{3}{@{}l}{\emph{ImageNet-100: generator and dataset-size comparisons}} \\
SRPO versus FLUX.1-dev & \textbf{+3.1} & [+1.6, +4.7] \\
FLUX.2-klein: adapter versus base-first & \textbf{+3.9} & [+2.2, +5.7] \\
SD-Turbo: adapter versus base-first & \textbf{+7.3} & [+5.3, +9.3] \\
Z-Image-Turbo: adapter versus base-first & \textbf{+2.4} & [+1.4, +3.4] \\
FLUX.2-klein: 100 adapted versus 3,000 released images/class & \textbf{+4.9} & [+2.2, +7.6] \\
\midrule
\multicolumn{3}{@{}l}{\emph{ImageNet-100: 500 real images/class, synthetic augmentation}} \\
SD 1.5: guidance 2 versus 7.5 & \textbf{+0.6} & [+0.1, +1.2] \\
klein: base-first versus released & \textbf{+2.0} & [+1.4, +2.7] \\
klein: released augmentation versus real-only & $-$0.1 & [$-$0.7, +0.5] \\
klein: base-first augmentation versus real-only & \textbf{+1.9} & [+1.2, +2.7] \\
klein: adapted versus released augmentation & \textbf{+1.8} & [+1.1, +2.5] \\
SD-Turbo: adapted versus released augmentation & \textbf{+2.9} & [+2.1, +3.6] \\
klein: adapted augmentation versus real-only & \textbf{+1.7} & [+0.9, +2.5] \\
SD-Turbo: adapted augmentation versus real-only & \textbf{+3.4} & [+2.6, +4.2] \\
\midrule
\multicolumn{3}{@{}l}{\emph{ImageNet-100: 50 real images/class, synthetic augmentation}} \\
klein: adapted versus released augmentation & \textbf{+12.1} & [+10.3, +14.1] \\
SD-Turbo: adapted versus released augmentation & \textbf{+13.1} & [+11.7, +14.6] \\
klein: adapted versus unguided-base augmentation & \textbf{+1.9} & [+0.8, +3.1] \\
SD-Turbo: adapted versus unguided-base augmentation & \textbf{+3.0} & [+1.8, +4.2] \\
\midrule
\multicolumn{3}{@{}l}{\emph{ImageNet-1k: full real-data training, klein augmentation}} \\
Adapted versus real-only: in distribution & $-$0.2 & [$-$0.4, 0.0] \\
Released versus real-only: in distribution & $-$0.7 & [$-$0.9, $-$0.5] \\
Adapted versus released: in distribution & \textbf{+0.5} & [+0.3, +0.7] \\
Adapted versus real-only: ImageNet-Sketch & \textbf{+3.3} & [+2.9, +3.7] \\
Adapted versus real-only: ImageNet-R & \textbf{+2.5} & [+2.0, +3.0] \\
Released versus real-only: ImageNet-Sketch & $-$0.4 & [$-$0.7, $-$0.1] \\
Released versus real-only: ImageNet-R & $-$0.5 & [$-$0.9, $-$0.2] \\
\bottomrule
\end{tabular}

  \caption{\textbf{Paired intervals for generator, data-scaling, and augmentation comparisons.}
  Accuracy differences in points [paired 95\% class-bootstrap CI] after averaging classifier runs.
  Original ImageNet-100 adapter/base-first comparisons pool three adapter-training runs; augmentation
  comparisons use one adapter run. Each set uses three classifier seeds, except the data-scaling
  comparison, which uses two. Augmentation comparisons match classifier-training steps and use the
  test set named in the row; otherwise evaluation is in distribution. Bold: interval above zero.
  These are existing estimates supporting \cref{sec:experiments,sec:supp-scaling,sec:supp-mixtures}.}
  \label{tab:supp-contrasts}

\end{minipage}\par\medskip

\FloatBarrier

\clearpage
\flushbottom
\twocolumn
\section*{LLM Usage Statement}
We used Claude Code (Opus 5.5) to assist with implementing experiment and analysis
code, and GPT-6.1-Sol to revise manuscript wording and presentation.
Qwen2.5-VL-7B-Instruct generated the experimental image captions and
class descriptions, as described in the supplementary material.
The authors reviewed the resulting code and text and take responsibility
for the reported results and conclusions.

{
    \small
    \bibliographystyle{ieeenat_fullname}
    \bibliography{main}
}

\end{document}